\pdfoutput=1

\documentclass[11pt]{article}

\usepackage[preprint]{acl}

\usepackage{times}
\usepackage{latexsym}

\usepackage[T1]{fontenc}

\usepackage[utf8]{inputenc}

\usepackage{microtype}

\usepackage{inconsolata}

\usepackage{graphicx}

\usepackage{booktabs, multirow} 

\usepackage{tcolorbox} 

\usepackage{hyperref}

\usepackage{longtable}

\usepackage[dvipsnames]{xcolor}

\title{NGQA: A Nutritional Graph Question Answering Benchmark\\for Personalized Health-aware Nutritional Reasoning}

\author{
\small
    Zheyuan Zhang\textsuperscript{1}\textsuperscript{*}, 
    Yiyang Li\textsuperscript{1}\textsuperscript{*}, 
    Nhi Ha Lan Le\textsuperscript{2}\textsuperscript{*}, 
    Zehong Wang\textsuperscript{1}, 
    Tianyi Ma\textsuperscript{1},
    Vincent Galassi\textsuperscript{1}, \\
\small
    \textbf{Keerthiram Murugesan \textsuperscript{3},
    Nuno Moniz\textsuperscript{1},
    Werner Geyer\textsuperscript{3},
    Nitesh V Chawla\textsuperscript{1}, 
    Chuxu Zhang\textsuperscript{4}
    Yanfang Ye\textsuperscript{1}\textsuperscript{$\dagger$}} \\
\small
    \textsuperscript{1}University of Notre Dame, 
    \textsuperscript{2}Brandeis University,
    \textsuperscript{3}IBM Research,
    \textsuperscript{4}University of Connecticut \\
\small
    \textsuperscript{*}Equal Contribution 
    \textsuperscript{$\dagger$}Corresponding Author
    \\
\small
    \texttt{\{zzhang42,yli62,zwang43,tma2,vgalassi,nmoniz2,nchawla,yye7\}@nd.edu},
    \\ 
\small
    \texttt{nhihlle@brandeis.edu}, 
    \texttt{keerthiram.murugesa@ibm.com}, \texttt{werner.geyer@us.ibm.com},
    \texttt{chuxu.zhang@uconn.edu} 
}

\begin{document}
\maketitle
\begin{abstract}
Diet plays a critical role in human health, yet tailoring dietary reasoning to individual health conditions remains a major challenge. Nutrition Question Answering (QA) has emerged as a popular method for addressing this problem. However, current research faces two critical limitations. On the one hand, the absence of datasets involving user-specific medical information severely limits \textit{personalization}. This challenge is further compounded by the wide variability in individual health needs. On the other hand, while large language models (LLMs), a popular solution for this task, demonstrate strong reasoning abilities, they struggle with the \textit{domain-specific} complexities of personalized healthy dietary reasoning, and existing benchmarks fail to capture these challenges. To address these gaps, we introduce the \textbf{N}utritional \textbf{G}raph \textbf{Q}uestion \textbf{A}nswering (\textbf{NGQA}) benchmark, the first graph question answering dataset designed for \textit{personalized nutritional health reasoning}. NGQA leverages data from the National Health and Nutrition Examination Survey (NHANES) and the Food and Nutrient Database for Dietary Studies (FNDDS) to evaluate whether a food is healthy for a specific user, supported by explanations of the key contributing nutrients. The benchmark incorporates three question complexity settings and evaluates reasoning across three downstream tasks. Extensive experiments with LLM backbones and baseline models demonstrate that the NGQA benchmark effectively challenges existing models. In sum, NGQA addresses a critical real-world problem while advancing GraphQA research with a novel domain-specific benchmark. Our codebase and dataset are available \href{https://anonymous.4open.science/r/NGQA-5E7F/README.md}{here}.
\end{abstract}

\begin{figure}[!t]
  \centering
  \includegraphics[width=\linewidth]{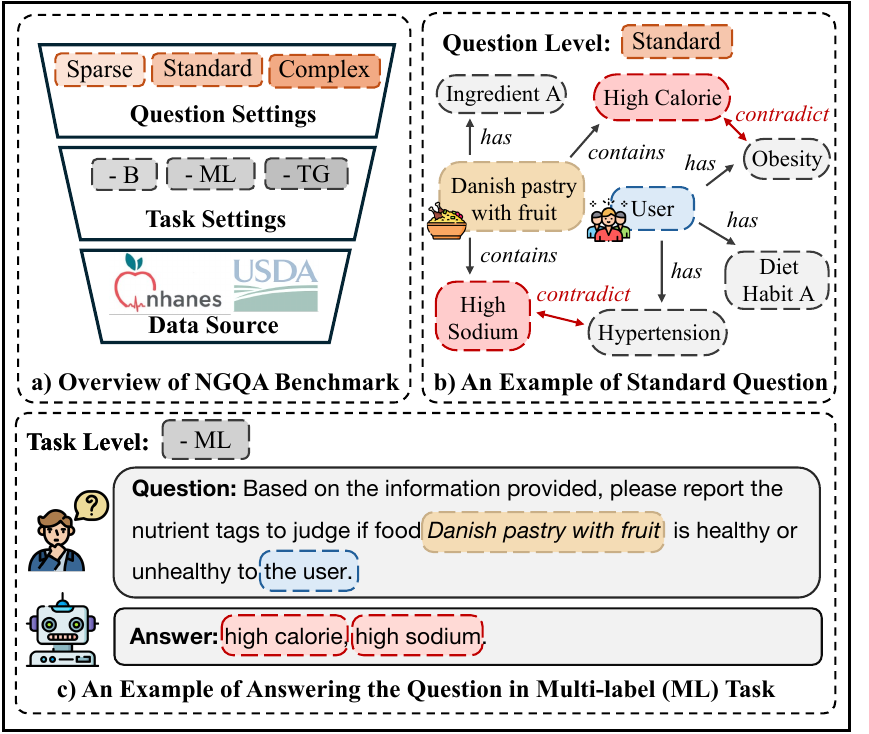}
  \vspace{-20pt}
  \caption{An Overview of NGQA Benchmark (a) along with a data showcase: (b) an example of the knowledge graph used for a standard level question and (c) the question and the answer of that question under the multi-label classification task (-ML) settings.}
  \vspace{-15pt}
  \label{fig:motivation}
\end{figure}

\section{Introduction}
Diet is a cornerstone of human health, playing a pivotal role in both maintaining well-being and preventing disease. Despite the well-documented benefits of balanced nutrition, unhealthy eating habits remain alarmingly prevalent in modern society \cite{WHO_2021_HealthyDiet}. In the United States alone, approximately 42.4\% of adults are classified as obese \cite{CDC_2020_Obesity}, and in 2017, poor dietary habits contributed to over 11 million deaths and a substantial number of disability-adjusted life-years (DALYs), often linked to factors such as excessive sodium intake \cite{Afshin_2019_DietaryRisks, WHO_2023_Obesity}. These statistics underscore an urgent need to promote healthier eating habits on a societal scale. However, nutritional health requires complex domain knowledge, and there is no one-size-fits-all solution for healthy diets, as the nutritional needs of individuals can vary widely based on their health conditions. For example, a diet suitable for someone with a high body mass index (BMI) may differ drastically from that of an individual with a low BMI. Likewise, while individuals recovering from opioid misuse may benefit from a high-protein diet, such dietary choices can be harmful to those managing chronic kidney disease \cite{mahboub2021nutritional}.

\textbf{Why this benchmark matters}: Numerous efforts have sought to address the challenges in personalized nutritional health, with Nutrition Question Answering (QA) emerging as a popular task \cite{min2022applications, bondevik2024systematic}. Recent advancements in large language models (LLMs) have demonstrated significant potential in this domain, offering sophisticated reasoning capabilities to analyze and interpret nutritional information \cite{mavromatis2024gnn}. However, these efforts remain constrained by two major limitations. First, to the best of our knowledge, no existing benchmark truly personalizes answers based on users’ specific health conditions, primarily due to the inaccessibility of individual medical data \cite{bolz2023hummus}. This lack of user-specific datasets has severely hindered the development of effective solutions. Second, while LLMs exhibit impressive reasoning capabilities in general domains, the medical and nutritional intricacies of this task impose severe limitations on their effectiveness \cite{mialon2023augmented}. Current benchmarks fail to capture the domain-specific complexities of personalized health-aware dietary reasoning, making it difficult to evaluate, let alone improve, these models in meaningful ways.

To address these critical gaps and advance the understanding of healthy diet personalization, we propose the \textbf{N}utritional \textbf{G}raph \textbf{Q}uestion \textbf{A}nswering (\textbf{NGQA}) benchmark. This is \textit{the first benchmark in the personalized nutritional health domain} to evaluate whether a specific food is healthy for a user, supported by detailed reasoning of the key contributing nutrients. By recognizing the intricate interplay between a user’s medical conditions, dietary behaviors, and the nutrition of foods, we frame this task as a knowledge graph question answering problem. Specifically, using data from the National Health and Nutrition Examination Survey (NHANES) and the Food and Nutrient Database for Dietary Studies (FNDDS), we construct the NGQA benchmark and categorize questions into three complexity settings: sparse, standard, and complex. Each question type is further evaluated through three downstream tasks, binary classification (-B), multi-label classification (-ML), and text generation (-TG), to explore distinct reasoning aspects (Figure-\ref{fig:motivation} (a)). We conduct extensive experiments using various LLM backbones and baseline models to ensure the benchmark is both appropriately challenging and meaningful for advancing the field. Our contributions can be summarized as follows:

\begin{itemize}
\item \textbf{Novel Benchmark for Personalized Nutrition.} We present NGQA, the first benchmark to incorporate users’ medical information in a nutritional question answering task, addressing a significant research gap in the domain of personalized healthy diet research.
\item \textbf{Advancing the GraphQA Ecosystem.} NGQA introduces a domain-specific benchmark and extends GraphQA benchmarks beyond datasets like \textit{WebQSP} and \textit{ExplaGraphs} in the general domain. This addition broadens the scope of GraphQA research, enabling a more comprehensive evaluation of GraphQA models’ capabilities beyond general reasoning tasks.
\item \textbf{Comprehensive Resource and Evaluation.} Through extensive experiments, NGQA provides a challenging benchmark, a complete codebase supporting the full pipeline from data preprocessing to model evaluation, and an extensibility for integrating new models. This comprehensive resource helps advance research in both personalized nutritional health and the broader GraphQA field.
\end{itemize}

\begin{figure*}[htbp!]
	\centering
	\includegraphics[width=1\linewidth]{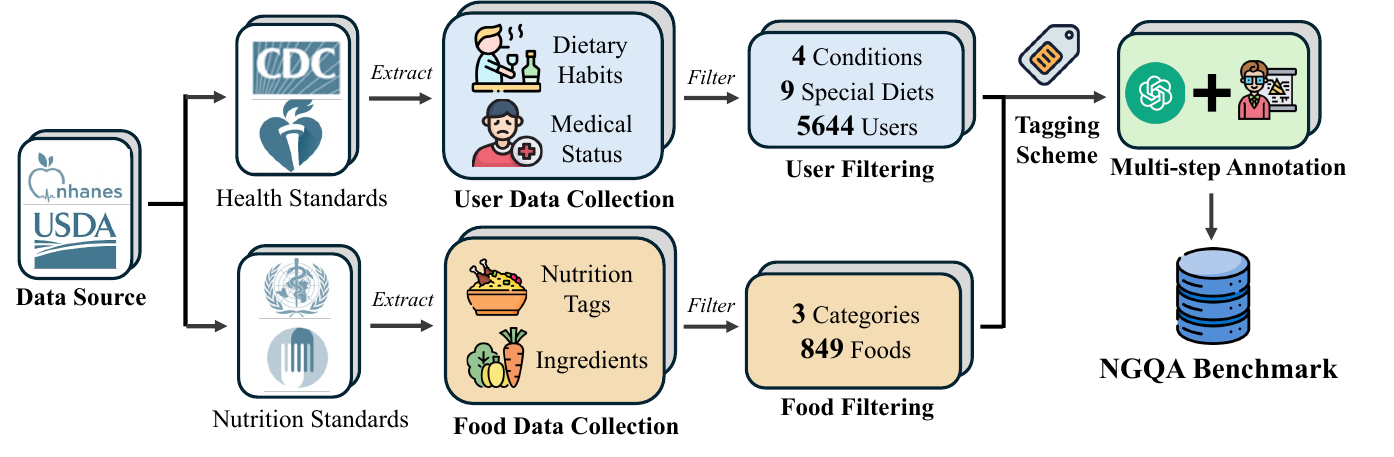}
        \vspace{-20pt}
	\caption{The NGQA benchmark construction process. Each stage shown in the figure is detailed in Section 3.For example, "User Data Collection" block, is introduced in Section 3.1 under the paragraph titled User Data Collection.}
        \vspace{-15pt}
    \label{fig:process}
\end{figure*}

\section{Related Work}

\noindent\textbf{Question Answering in Nutritional Health Domain.}
Question answering has become an essential tool in the nutritional and health domain, offering a flexible framework for applications such as food recommendation \cite{min2022applications, bondevik2024systematic}. Knowledge graphs (KGs) have been widely used to model relationships between foods, ingredients, and health, supporting tasks like ingredient substitution and adaptive dietary recommendations \cite{haussmann2019foodkg, chen2021personalized, fatemi2023learning, xu2024adaptive}. Recent approaches incorporate health metrics into QA systems, focusing on recipe recommendations and nutritional ontologies \cite{li2023health, seneviratne2021personal}. However, existing methods lack true personalization, as highlighted by \cite{bolz2023hummus}, due to the absence of user-specific medical data. Our work fills this gap by introducing the first GraphQA benchmark for personalized nutritional health, enabling models to provide tailored nutritional reasoning and explanations.

\noindent\textbf{Graph Retrieval Augmented Generation.}
Knowledge Graph Question Answering (KGQA) has progressed from early semantic parsing and retrieval-based methods to advanced techniques leveraging large language models (LLMs) and graph neural networks (GNNs) for reasoning and retrieval \cite{jiang2023structgpt, kim2023kg, gao2024two}. Building on this progress, Graph-Retrieval Augmented Generation (Graph-RAG) has emerged as a widely studied method, offering more precise, context- and structure-aware reasoning compared to traditional text-based RAG methods \cite{lewis2020retrieval, lazaridou2022internet, guo2024knowledgenavigator, wen2023mindmap}. Despite the development of various LLM-powered models, benchmarks for the Graph-RAG task remain scarce and lack standardization. Early benchmarks focus primarily on general graph tasks such as shortest paths and node degree \cite{fatemi2023talk, wang2024can}, while \cite{he2024g} introduces a GraphQA benchmark for complex reasoning using general-purpose datasets. Building on their framework, we develop the first domain-specific benchmark in the nutritional health domain, bridging the gap between general GraphQA research and personalized health-aware reasoning. More detailed literature is available in Appendix-\ref{appendix:Additional Related Work}.

\section{NGQA Benchmark}
\subsection{Data Collection}
\textbf{Data Source.} Using data from the National Health and Nutrition Examination Survey (NHANES) and the Food and Nutrient Database for Dietary Studies (FNDDS), we construct the first GraphQA benchmark designed to address personalized healthy nutrition intake questions. This benchmark integrates detailed user health profiles, dietary behaviors, and comprehensive food nutritional information, enabling a fine-grained analysis of how individual health conditions interact with food nutrition. By representing these relationships through graph structures, the benchmark supports answering complex nutritional questions while capturing the intricate interplay between users’ medical conditions and dietary choices. The following sections provide a detailed discussion of these datasets and their integration into our benchmark.

\textbf{User Data Collection.}
The NHANES dataset forms the foundation of our work for collecting user data. We extract medical information, dietary habits, and food intake records to construct the graph. Specifically, NHANES provides laboratory reports detailing body metrics like Body Mass Index (BMI) and blood pressure, along with biochemical markers such as blood urea nitrogen. It also includes questionnaire responses on prescription drug usage, adherence to special diets, and overall health status. Additionally, NHANES records users’ food intake history and dietary behaviors, such as the frequency of adding salt at the table. Our study incorporates 54 distinct dietary habits, with detailed data processing methods provided in Appendix-\ref{appendix:Benchmark Details}. This comprehensive dataset serves as the backbone of our graph, capturing user health conditions and dietary patterns with granular detail.

\textbf{Food Data Collection.}
Nutritional information for food items is sourced from FNDDS. FNDDS connects NHANES food codes to detailed nutritional data cataloged in the What We Eat in America (WWEIA) database. Using FNDDS, we associate each food item in NHANES with its full nutritional composition. Additionally, FNDDS links food items to ingredient information and classifies them into broader food categories. For example, a food item like "apple" is linked to its nutrient values (e.g., sugars, vitamins) and assigned to the category "fruits." These associations enrich the graph by providing node-level data for food, ingredients, and categories.

\textbf{Tagging Scheme.} To evaluate whether a food is specifically healthy for a user based on their personal health conditions, we propose a tagging scheme that assigns nutrition-related tags to both users and foods. This systematic framework aligns food nutritional properties with user health needs, enabling robust assessments of food suitability.

For food tagging, we build upon established guidelines and introduce newly applied standards. Prior works have utilized recommendations from the World Health Organization (WHO) and the Food Standards Agency (FSA) \cite{wang2021market2dish}, while we extend this by incorporating the more detailed EU Nutrition \& Health Claims Regulation \cite{EC2006} and the Codex Alimentarius Commission (CAC) \cite{FAO1985, FAO1997}. These standards define precise thresholds for nutrient claims. For instance, the EU regulation permits labeling a food as "low sodium" only if it contains no more than 0.12 g of sodium per 100 g \cite{EC2006}. Foods meeting such criteria are tagged with corresponding labels like "low\_sodium" or "high\_protein", reflecting their nutritional properties.

On the user side, health tags are derived from the NHANES dataset, which includes laboratory results and self-reported health information. For example, users with high blood pressure, as defined by American Heart Association (AHA) thresholds or similar guidelines, are tagged with "hypertension," indicating that a low-sodium diet would be beneficial \cite{grillo2019sodium, smyth2014sodium}. 

By linking health and food tags, our scheme effectively represents personalized dietary needs and captures the interplay between medical conditions and nutritional requirements. The detailed standards and additional tags for other nutrients and health conditions are described in Appendix-\ref{appendix:Benchmark Details}. By integrating this methodology into our graph-based benchmark, we provide a framework for advancing personalized dietary reasoning and evaluating models in this domain.

\subsection{Data Annotation}
Real-world data is inherently messy and incomplete, and the datasets we use are no exception. Spanning from 2003 to 2020, NHANES provides data for approximately 100,000 users and over 2 million food records. While this dataset offers an invaluable resource for studying nutrition and health, it includes inconsistencies, ambiguities, and irrelevant entries. To establish a scientifically robust and meaningful benchmark, precise data annotation is essential. This involves not only cleaning and filtering the data but also carefully defining and validating annotations to accurately capture real-world relationships between health conditions, dietary behaviors, and food options. Our annotation process refines both user and food datasets to ensure relevance, accuracy, and applicability to real-life scenarios.

\begin{figure*}[htbp!]
	\centering
	\includegraphics[width=1\linewidth]{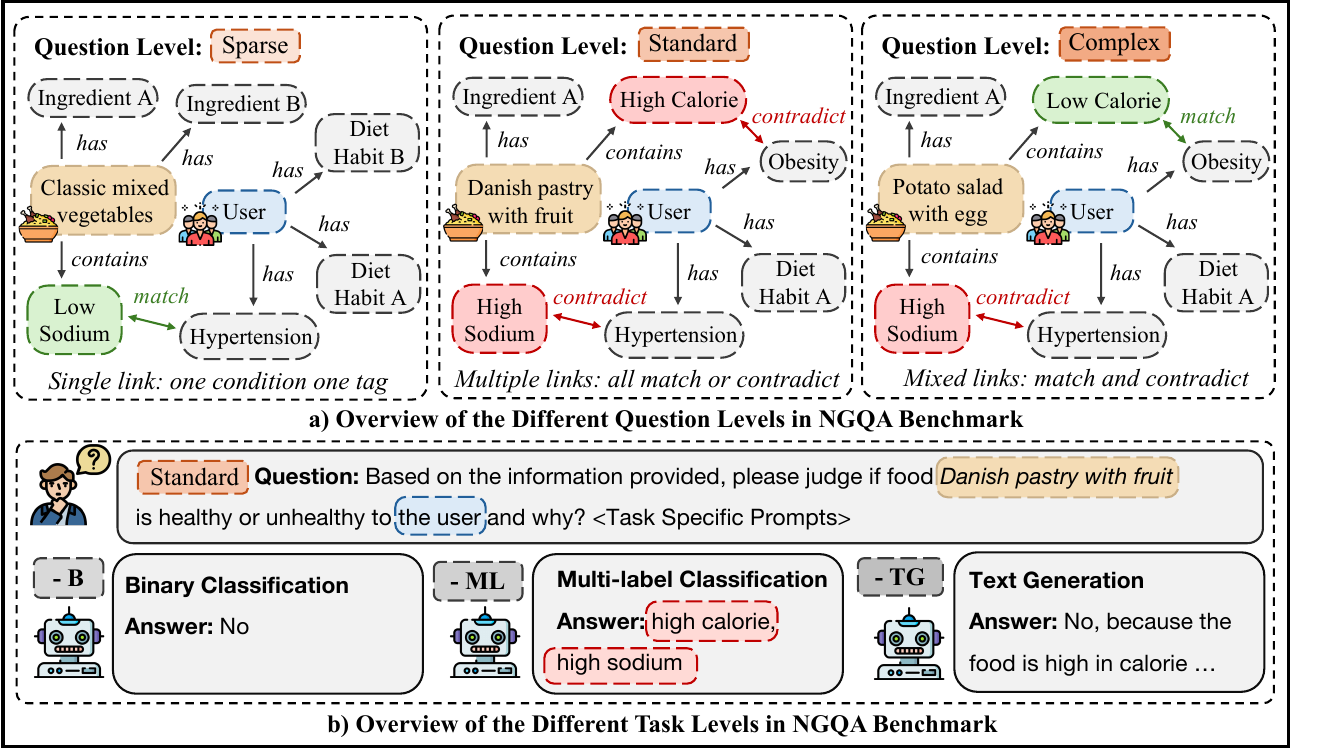}
        \vspace{-20pt}
	\caption{The illustration of different question levels and task levels.}
        \vspace{-15pt}
    \label{fig:task}
\end{figure*}

\textbf{User Filtering.} Annotating user data requires careful consideration of the complex interactions between nutrition and health. For instance, elevated blood urea nitrogen (BUN) levels may indicate kidney dysfunction, warranting a low-protein diet, but could also result from insufficient water intake. To maintain scientific rigor and practical relevance, we focus on annotating four prevalent health statuses—obesity, hypertension, opioid misuse, and diabetes—that are directly influenced by dietary interventions. Additionally, we annotate nine special diets reported by users, reflecting health-related dietary practices. Further details on the definitions and implications of these health statuses and diets are provided in the Appendix-\ref{appendix:Benchmark Details}. To ensure consistency and relevance, we exclude users under 18, focusing solely on adult dietary patterns.

\textbf{Food Filtering.} For food annotation, we identify practical entries in the FNDDS database that align with real-world dietary reasoning. While FNDDS supports comprehensive nutritional analysis, it includes many entries unsuitable for practical use, such as raw ingredients or standalone additives. To address this, we restrict our focus to the "mixed dishes" category, as it represents combined recipes closest to real-life diets. Additionally, we include other relevant categories, such as bakery products and desserts (definitions of FNDDS categories are available in the Appendix-\ref{appendix:Standards and Regulation}). Finally, we apply a keyword-based deduplication method to remove highly similar entries.

\textbf{Multi-step Annotation.} Using the previously defined standards and tagging schemes, our annotation process systematically establishes "match" or "contradict" relationships between user health conditions and food nutritional profiles. For example, the tag "high\_calorie" contradicts the condition "obesity", while "low\_sodium" matches with "hypertension". To ensure accuracy and reliability, we adopt a multi-step annotation process. After initial filtering and tagging, large language models (LLMs) perform an initial sanity check to identify inconsistencies or anomalies in the annotations. Subsequently, three human annotators with domain expertise review and cross-validate the results to eliminate remaining inaccuracies. By combining automated checks with human validation, our rigorous annotation strategy captures the real-life complexities of personalized nutrition while maintaining high standards of quality and reliability. 

\section{Task Definition and Evaluation}
\subsection{Question Setting}
With the annotated data in place, we designed three distinct types of questions, i.e., \textit{sparse}, \textit{standard}, and \textit{complex}, to capture varying levels of difficulty and emulate real-world scenarios in personalized nutrition reasoning. This stratification ensures that our benchmark accommodates a wide range of research and application needs, spanning from controlled, idealized setups to challenging, real-life cases, as illustrated in Figure-\ref{fig:task} (a).

\textbf{Sparse questions} address scenarios with minimal available information. In this setting, each food has only one nutrition tag linked to a single user health condition. This setup reflects real-world cases where labels are scarce or data is incomplete, challenging models to reason effectively with limited information. Although sparse questions may appear simple to human observers, the unique link between the user and the food significantly increases the difficulty of subgraph retrieval, making models vulnerable to interference from irrelevant nodes.

\textbf{Standard questions} represent the balanced and idealized scenarios in our benchmark. In this category, foods are linked to multiple nutrition tags, which either match or contradict several user health conditions. This configuration reflects controlled cases where the relationship between dietary choices and health outcomes is clear-cut, enabling a focused evaluation of model performance. Standard questions serve as a foundation for benchmarking in structured and well-defined environments.

\textbf{Complex questions} are designed to replicate the intricacies of real-life nutritional decision-making. Foods in this category may simultaneously have tags that both match with and contradict a user’s health conditions. For instance, a food may be low in sodium (beneficial for hypertension) but also high in sugar (problematic for diabetes). These scenarios require models to navigate conflicting information, prioritize user health needs, and perform nuanced trade-off reasoning. This category closely mirrors the ambiguous and multifaceted challenges of real-world dietary decisions. 

The benchmark’s statistical breakdown is presented in Table-\ref{tab:benchmark_stats}. To further evaluate the complexity and informativeness of the questions, we introduce the Signal-to-Noise Ratio (SNR). SNR measures the ratio of nodes or tags relevant to the answer (\textit{signal}) against the total nodes or tags in the graph (\textit{noise}). As shown in Table-\ref{tab:snr_stats}, sparse questions exhibit the lowest SNR, reflecting the limited resources available for these tasks. Conversely, complex questions, despite containing conflicting information, achieve the highest SNR, underscoring the rich contextual information necessary for accurate reasoning. More statistics of the benchmark are available in Appendix-\ref{appendix:Additional Statistics}.

\subsection{Task Setting}

To enhance the generality and versatility of our benchmark, we design three distinct downstream task types, each centered on the same domain question but requiring different forms of output, as illustrated in Figure-\ref{fig:task} (b). This diversity ensures the benchmark accommodates a wide range of methodologies and research focuses while fostering innovation in addressing personalized nutrition challenges. The tasks are defined as follows:

\begin{table}[!t]
    \centering
    \setlength{\tabcolsep}{6pt} 
    \resizebox{\columnwidth}{!}{ 
        \begin{tabular}{lccc}
            \toprule
            \textbf{Question Level} & \textbf{\# Records} & \textbf{Avg. \# Nodes} & \textbf{Avg. \# Edges} \\
            \midrule\midrule
            Sparse      & 8,490 & 25.84 & 24.86  \\
            Standard    & 3,622 & 28.16 & 28.98  \\
            Complex     & 1,690 & 30.94 & 34.04  \\
            \bottomrule
        \end{tabular}
    }
    \vspace{-5pt}
    \caption{Statistics of the Benchmark by Question Level.}
    \vspace{-10pt}
    \label{tab:benchmark_stats}
\end{table}

\begin{table}[!t]
    \centering
    \setlength{\tabcolsep}{6pt} 
    \resizebox{\columnwidth}{!}{ 
        \begin{tabular}{lcc}
            \toprule
            \textbf{Question Level} & \textbf{Avg. Node SNR} & \textbf{Avg. Tag SNR} \\
            \midrule\midrule
            Sparse      & 16.37 & 19.30 \\
            Standard    & 24.68 & 49.39 \\
            Complex     & 31.57 & 76.32 \\
            \bottomrule
        \end{tabular}
    }
    \vspace{-5pt}
    \caption{Signal-to-Noise Ratio (SNR) by Question Level.}
    \vspace{-10pt}
    \label{tab:snr_stats}
\end{table}

\textbf{Binary Classification (-B)}: This task requires a simple "yes" or "no"  response, indicating whether a specific food is suitable for a user based on their health profile. It emphasizes straightforward decision-making, reflecting applications like automated diet advisories or recommendation systems.

\textbf{Multi-Label Classification (-ML)}: In this task, models must retrieve the nutritional tags associated with a food and determine which match with or contradict the user’s health conditions. By demanding richer output, this task evaluates the model’s ability to leverage graph information and identify nuanced relationships.

\textbf{Text Generation (-TG)}: The output is a natural language explanation of why a food is healthy or unhealthy for a user. This task assesses a model’s capability for interpretable and user-friendly reasoning, which is crucial for real-world applications such as personalized dietary assistant chatbots.

\subsection{Evaluation Metrics}

To evaluate model performance, we adopt task-specific metrics tailored to each type. For classification tasks, we use standard metrics like accuracy, recall, precision, and F1 score for comprehensive performance assessment. Multi-label classification tasks extend these metrics to their weighted versions, accounting for the distribution of multiple labels across samples. Text generation tasks are evaluated with widely used metrics such as ROUGE, BLEU, and BERT scores, which collectively assess relevance and semantic similarity to reference texts. The definition of ground truths is available in Appendix-\ref{appendix:Benchmark Details}. This multifaceted design supports diverse model architectures and evaluation strategies, providing a robust foundation for advancing personalized nutrition research. By bridging the gap between controlled research environments and the complexities of real-world applications, our benchmark fosters innovation and opens new avenues for addressing healthy dietary reasoning.

\section{Experiments}

\begin{table*}[t]
\centering
\resizebox{\textwidth}{!}{
\begin{tabular}{cc|cccc|cccc|ccccc}
\toprule
\multirow{2}{*}{\textbf{Question Level}} & \multirow{2}{*}{\textbf{Method}} & \multicolumn{4}{c}{\textbf{a) Binary Classification (-B)}} & \multicolumn{4}{c}{\textbf{b) Multi-label Classification (-ML)}} & \multicolumn{5}{c}{\textbf{c) Text Generation (-TG)}} \\ 
\cmidrule{3-6} \cmidrule{7-10} \cmidrule{11-15}
 & & \textbf{Accuracy} & \textbf{Recall} & \textbf{Precision} & \textbf{F1} & \textbf{Accuracy} & \textbf{Recall} & \textbf{Precision} & \textbf{F1} & \textbf{ROUGE-1} & \textbf{ROUGE-2} & \textbf{ROUGE-L} & \textbf{BLEU} & \textbf{BERT} \\\midrule
\multirow{5}{*}{\textbf{Sparse}} & \textbf{Plain} & 0.5973 & 0.1634 & \textbf{1.0000} & 0.2810 & 0.1798 & 0.9943 & 0.2109 & 0.3442 & 0.5385 & 0.4775 & 0.5385 & 0.2838 & 0.9370 \\ 
& \textbf{KAPING} & 0.5347 & 0.0541 & 0.7246 & 0.1006 & 0.1753 & 0.9915 & 0.2075 & 0.3394 & 0.5234 & 0.4600 & 0.5234 & 0.2674 & 0.9353 \\ 
& \textbf{CoT-Zero} & 0.6604 & 0.2951 & 0.9983 & 0.4555 & 0.2032 & 0.9958 & 0.2435 & 0.3842 & 0.5463 & 0.4842 & 0.5462 & 0.2889 & 0.9388 \\ 
& \textbf{CoT-BAG} & 0.6038 & 0.1769 & \textbf{1.0000} & 0.3006 & 0.2134 & \textbf{0.9966} & 0.2520 & 0.3945 & 0.5481 & 0.4886 & 0.5480 & 0.2930 & 0.9385 \\ 
& \textbf{ToG} & \textbf{0.7729} & \textbf{0.5383} & 0.9817 & \textbf{0.6953} & \textbf{0.2439} & 0.9128 & \textbf{0.2986} & \textbf{0.4333} & \textbf{0.6254} & \textbf{0.5710} & \textbf{0.6251} & \textbf{0.3612} & \textbf{0.9465} \\ \midrule
\multirow{5}{*}{\textbf{Standard}} & \textbf{Plain} & 0.5762 & 0.1989 & \textbf{1.0000} & 0.3317 & 0.4909 & 0.9980 & 0.4901 & 0.6528 & 0.7219 & 0.6321 & 0.6941 & 0.4840 & 0.9618 \\ 
& \textbf{KAPING} & 0.5022 & 0.0637 & 0.9313 & 0.1192 & 0.4593 & 0.9956 & 0.4624 & 0.6272 & 0.7087 & 0.6237 & 0.6764 & 0.4617 & 0.9599 \\ 
& \textbf{CoT-Zero} & 0.6565 & 0.3507 & \textbf{1.0000} & 0.5193 & 0.5390 & 0.9967 & 0.5447 & 0.6963 & 0.7329 & 0.6443 & 0.7049 & 0.4939 & 0.9630 \\ 
& \textbf{CoT-BAG} & 0.5900 & 0.2249 & \textbf{1.0000} & 0.3673 & 0.5599 & \textbf{0.9982} & 0.5611 & 0.7091 & 0.7333 & 0.6456 & 0.7032 & 0.4951 & 0.9630 \\ 
& \textbf{ToG} & \textbf{0.8628} & \textbf{0.7411} & 0.9993 & \textbf{0.8511} & \textbf{0.6189} & 0.8843 & \textbf{0.6793} & \textbf{0.7464} & \textbf{0.8182} & \textbf{0.7632} & \textbf{0.7817} & \textbf{0.6112} & \textbf{0.9716} \\ \midrule
\multirow{5}{*}{\textbf{Complex}} & \textbf{Plain} & 0.6598 & 0.0636 & 0.9750 & 0.1194 & 0.7185 & 0.9721 & 0.7374 & 0.8358 & 0.7356 & 0.6510 & 0.7001 & 0.4949 & 0.9599 \\ 
& \textbf{KAPING} & 0.6574 & 0.0571 & 0.9722 & 0.1079 & 0.6883 & \textbf{0.9758} & 0.7129 & 0.8093 & 0.7394 & 0.6634 & 0.7016 & 0.4839 & 0.9602 \\ 
& \textbf{CoT-Zero} & 0.6627 & 0.0718 & 0.9778 & 0.1337 & 0.7453 & 0.9735 & 0.7679 & 0.8557 & 0.7478 & 0.6599 & 0.7103 & 0.5048 & 0.9615 \\ 
& \textbf{CoT-BAG} & 0.6627 & 0.0701 & \textbf{1.0000} & 0.1311 & \textbf{0.7546} & 0.9631 & 0.7801 & \textbf{0.8587} & 0.7467 & 0.6622 & 0.7080 & 0.5049 & 0.9611 \\ 
& \textbf{ToG} & \textbf{0.7473} & \textbf{0.3964} & 0.8100 & \textbf{0.5323} & 0.6153 & 0.6989 & \textbf{0.8119} & 0.7303 & \textbf{0.7729} & \textbf{0.6915} & \textbf{0.7366} & \textbf{0.5313} & \textbf{0.9639} \\ 
\bottomrule
\end{tabular}}
\vspace{-5pt}
\caption{Experimental results based on five baseline methods on the three tasks with the three question levels using the GPT-4o-mini. The best performance of each group is bolded.}
\vspace{-15pt}
\label{tab:experiment-results}
\end{table*}

\subsection{Experiment Settings}
In this section, we conduct extensive experiments to evaluate existing Graph-RAG models' reasoning capability on the proposed benchmark. For baseline models, we select the five most classical baselines: KAPING \cite{baek2023knowledge}, CoT-Zero \cite{kojima2022large}, CoT-BAG \cite{wang2024can}, ToG \cite{sunthink}, and a naive plain Graph-RAG pipeline (implementation details in Appendix-\ref{appendix:Implementation Details}). For the main experiments, we choose GPT-4o-mini as the LLM backbone, we also conduct additional experiments on a series of other classic LLM backbones in Appendix-\ref{appendix:Additional Experiments}. Note that we didn't select the most advanced LLM backbones or the most sophisticated fine-tuned baselines because we argue our contributions focus primarily on the proposed benchmark with the novel tasks for this specific domain, and the experiment results along with the hallucination analyses have demonstrated our tasks are properly designed where the classic baselines can be adequately challenged while maintaining efficiency. In the following sections, we go through the experiment results for each task. 

\subsection{Binary Classification Task}
Table-\ref{tab:experiment-results} (a) presents the performance of baseline models on the binary classification task, which evaluates the models’ ability to provide a decisive "yes" or "no" response based on summarized reasoning. The results reveal a notable conservatism in model behavior, as evidenced by the low recall scores. This likely stems from the sensitive nature of medical questions, where LLMs try to avoid offering simple "yes" answers without explanations unless their confidence is exceptionally high. Despite this challenge, the experiments yield two important insights into how external domain knowledge can support LLMs in this scenario. First, increasing the number of links in the graph (e.g., from Sparse to Standard questions) consistently improves recall across all baselines. This indicates that richer external knowledge provides LLMs with greater context and reassurance, enabling them to produce more confident positive answers. Second, ToG significantly outperforms other baselines, showing performance gains unique to this task. We attribute this improvement to ToG’s effective pruning mechanism, which removes irrelevant nodes and increases the SNR. By reducing noise and focusing on relevant information, ToG enhances LLMs’ ability to make confident and accurate decisions.

\begin{figure}[t!]
  \centering
  \includegraphics[width=\linewidth]{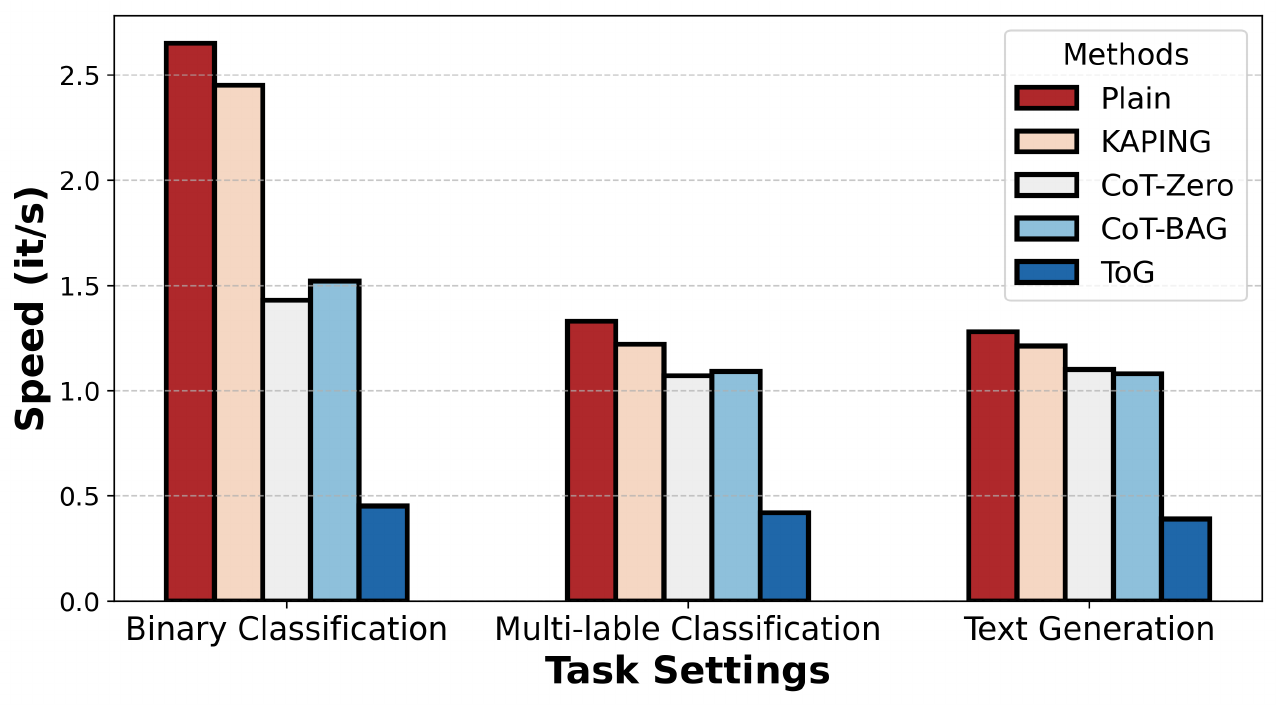}
  \vspace{-20pt}
  \caption{Efficiency analysis of the five baseline methods across three tasks.}
  \vspace{-15pt}
  \label{fig:runtime}
\end{figure}

\begin{figure}[htbp!]
  \centering
  \includegraphics[width=\linewidth]{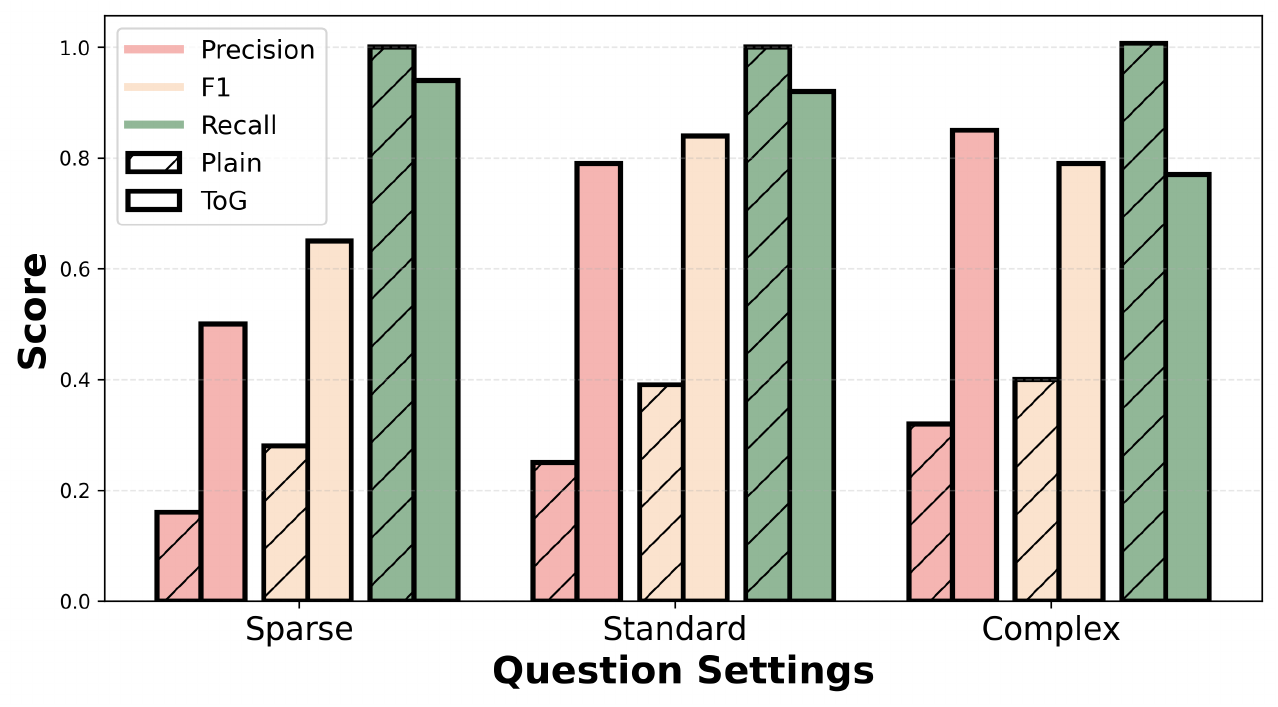}
  \vspace{-20pt}
  \caption{Retrieval quality of ToG vs. Plain across three types of questions on recall, precision and F1.}
  \vspace{-15pt}
  \label{fig:retrieval}
\end{figure}

\subsection{Multi-label and Text Generation Task}
Table-\ref{tab:experiment-results} (b) and (c) present the performance of baseline models on the multi-label classification (ML) and text generation (TG) tasks. The ML task evaluates models’ ability to retrieve nutrition tags associated with foods and user health conditions, while the TG task tests their capacity to generate natural language explanations, offering a more comprehensive and realistic evaluation. The results reveal similar patterns across tasks: while baselines are competent at identifying nutrition tags from the graph, the primary challenge lies in correctly identifying the relevant tags based on user health conditions, as indicated by the overall high recall scores in the ML task.

Both tasks are most challenging on sparse question sets due to their low-resource nature. Conversely, models achieve the best performance on complex question sets, which may appear counterintuitive. However, as shown in Table-\ref{tab:snr_stats}, complex questions have a higher Signal-to-Noise Ratio (SNR), providing models with a clearer signal that offsets their logical complexity. Additionally, the ToG model performs similarly on the standard and complex question sets due to its pruning process, which increases SNR by removing irrelevant nodes. While effective, this process can also discard valuable information, leading to lower performance on complex questions. This trade-off contrasts with ToG’s success in binary classification task and highlights the comprehensiveness of our benchmark, which challenges models across diverse scenarios to uncover their strengths and weaknesses.

\subsection{Efficiency and Retrieval Quality}
Beyond model performance, efficiency is a critical consideration in Graph-RAG systems. To evaluate this, we conduct an efficiency analysis of baseline models on our benchmark, as shown in Figure-\ref{fig:runtime}. As can be seen, the binary classification task exhibits the fastest runtime, as it requires the shortest output. In contrast, the multi-label classification and text generation tasks involve longer outputs, leading to slower performance. Due to ToG’s reliance on multiple LLM calls during the retrieval process, its runtime is significantly slower compared to other methods. Additionally, the quality of subgraph retrieval plays a crucial role in downstream reasoning. To assess this, we perform a retrieval quality analysis using ToG as a case study, comparing it against a plain Graph-RAG pipeline, as illustrated in Figure-\ref{fig:retrieval}. As shown, the retrieval scores of ToG align with its performance in the main experiments, confirming our assumption that fluctuations in ToG’s performance are rooted in its pruning process during the subgraph retrieval phase.

\begin{figure}[t!]
  \centering
  \includegraphics[width=\linewidth]{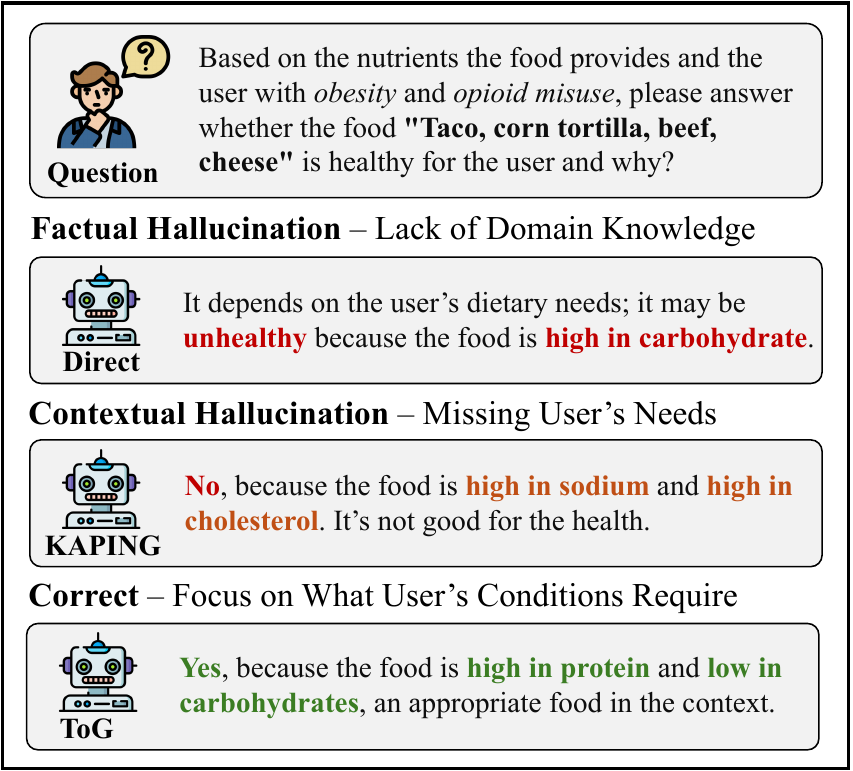}
  \vspace{-20pt}
  \caption{A case study of error analysis.}
  \vspace{-20pt}
  \label{fig:case}
\end{figure}

\subsection{Error Analysis}
In this section, we analyze the types of hallucinations observed in our experiments using a specific example and demonstrate the importance of external domain knowledge in mitigating these errors.

Traditional LLM-enhanced methods are well-known for their susceptibility to hallucination errors, particularly in domain-specific tasks like nutritional health \cite{mialon2023augmented}. Figure-\ref{fig:case} illustrates an example where we evaluate whether the food "Taco, corn tortilla, beef, cheese" is a healthy option for a user who is obese and recovering from opioid misuse. Our analysis identifies two main types of hallucinations. The first is \textit{Factual Hallucination}, where the model produces incorrect or irrelevant information, often due to reliance on general knowledge not explicitly included in the graph. These errors are common when LLMs perform direct inference without external knowledge and occasionally occur when retrieved graphs contain noise. For example, the model incorrectly deemed the taco unsuitable, overlooking the fact that corn tortillas are relatively low in carbohydrates. 

The second type is \textit{Contextual Hallucination}, where the model fails to prioritize tags that directly relate to the user’s health profile, focusing instead on less relevant attributes. This issue is less pronounced in ToG due to its ability to retrieve compact, focused subgraphs, unlike simpler methods like KAPING and CoT-Zero, which lack effective pruning. In this case, the taco’s high sodium and cholesterol overshadowed its alignment with the user’s specific health needs for a low-carb, high-protein diet, leading to a less optimal assessment.

In summary, these hallucinations highlight the importance of our domain-specific benchmark in establishing a rigorous framework to evaluate and improve LLMs, advancing both the nutritional health domain and Graph-RAG research while fostering the development of more robust and generalizable models (More examples in Appendix-\ref{appendix:Error Analysis}).

\section{Conclusion}

In this work, we introduce the Nutritional Graph Question Answering (NGQA) benchmark, the first dataset designed to address the critical challenges of personalized nutritional health reasoning. By leveraging user-specific medical data and framing the problem as a knowledge graph question answering task, NGQA bridges the gap between general-purpose benchmarks and domain-specific applications. Our benchmark not only advances the scope of GraphQA research by incorporating complex, real-world nutritional scenarios but also provides a comprehensive resource for evaluating and improving models in this domain. We believe NGQA lays the foundation for future research in personalized diet and health-aware reasoning, fostering innovation in both nutritional health and GraphQA.

\newpage
\section*{Limitation}
In this section, we discuss the limitations of this work and outline directions for future research. First, the benchmark includes a limited number of health conditions, though more are available. For example, osteoporosis suggests a high-calcium diet, a renal diet indicates low protein intake, and high low-density lipoprotein (LDL) levels may call for a low-cholesterol diet. As noted in the paper, we prioritized conditions most prevalent in the United States and most relevant to dietary interventions, but expanding to include additional conditions could enhance coverage and utility. Second, while we focus on the interplay between dietary behaviors and medical conditions, other factors, such as food insecurity, remain unexplored. NHANES offers extensive socioeconomic data, presenting opportunities to extend the benchmark to account for broader determinants of dietary decision-making. Third, for simplicity, complex questions are reduced to binary classification by counting "match" and "contradict" tags. However, real-life dietary decisions require nuanced trade-offs and reasoning that go beyond this approach. More sophisticated evaluation methods could better reflect practical scenarios. Lastly, the benchmark could benefit from additional tasks. For example, the existing graphs support questions like, "What alternative foods could meet a user’s dietary preferences and medical needs?" Incorporating such tasks would broaden the benchmark’s scope and encourage further innovation. Despite these limitations, this work establishes a robust baseline as a pioneering effort in personalized nutrition reasoning. We defer these challenges to future work, envisioning the benchmark as a foundation for ongoing advancements in this critical domain.

\section*{Ethics and Privacy Statement}
Safeguarding privacy and adhering to ethical principles are paramount when working with sensitive health-related data. The National Health and Nutrition Examination Survey (NHANES) serves as a benchmark in this regard, strictly complying with confidentiality protocols mandated by public legislation. These robust privacy measures enable us to achieve our research goals while remaining fully aligned with the survey’s established guidelines. Notably, the NHANES dataset is anonymized, with personally identifiable information (PII)—such as social security numbers and physical addresses—removed. Despite the absence of PII, the dataset retains its utility for detailed analyses, allowing us to investigate the relationship between users’ medical data and health-aware food recommendations as presented in this study. Additionally, in practical applications, the generated recommendations and interpretations are treated as personal medical records, ensuring sustained privacy protection. By adhering to these principles, our research maintains the highest levels of ethical responsibility and data privacy.

\bibliography{reference}

\newpage
\appendix
\section{Additional Related Work}
\label{appendix:Additional Related Work}

\subsection{Prior Works in Nutrition Personalization}
With growing awareness of the importance of dietary health, various studies have sought to incorporate health metrics into applications such as food recommendation systems. These approaches can be grouped into three primary categories. First, some research emphasizes single indicators like calorie or fat content, as highlighted in works by Ge et al. \cite{ge2015health} and Shirai et al. \cite{shirai2021identifying, li2024cheffusion}, though such metrics often fail to represent the multifaceted nature of a balanced diet. Second, simulated health data has been utilized, as demonstrated by Wang et al. \cite{wang2021market2dish}, but these methods often diverge from real-world data distributions. Finally, recent studies have applied global health guidelines to develop composite health scores, such as those by Bolz et al. \cite{bolz2023hummus} and Zhang et al. \cite{zhang2024greenrec}. However, foods deemed healthy by general standards can still negatively affect certain individuals \cite{yue2021overview}, highlighting the absence of a universal solution. The primary challenge remains the scarcity of accurate user health data, a gap our benchmark uniquely addresses.

\subsection{Knowledge Graph Question Answering}
Knowledge Graph Question Answering (KGQA) has undergone significant advancements, evolving from early approaches such as semantic parsing and retrieval-based methods. Initial models translated natural language queries into structured formats like SPARQL for execution on knowledge graphs \cite{sun2019pullnet, zhang2022subgraph}. Many of these methods employed pre-trained models like BERT for query encoding and used frameworks such as GNNs or LSTMs for retrieving entities and subgraphs \cite{yasunaga2021qa, taunk2023grapeqa}.

More recent progress integrates large language models (LLMs) to improve both retrieval efficiency and reasoning ability \cite{sanchez2022effects, liu2024towards, tan2024democratizing}. Approaches like Jiang et al. \cite{jiang2023structgpt} and Wang et al. \cite{wang2023knowledgpt} utilize LLMs to transform queries into formats such as SQL or SPARQL, enhancing retrieval accuracy. Others, such as Kim et al. \cite{kim2023kg} and Gao et al. \cite{gao2024two}, focus on reasoning over retrieved subgraphs or triples, tackling multi-hop reasoning tasks in KGQA. However, most benchmarks in this field are designed for general-purpose datasets and fail to address domain-specific complexities, such as the challenges unique to nutritional health reasoning.

\subsection{Graph-Retrieval Augmented Generation}
Graph neural networks exhibit powerful potentials in dealing with complicated structural data \cite{wang2024subgraph, liu2023fair, wang2024gft} and it can facilitate LLM to better understand real world tasks \cite{wang2024canllm, huang2024application, liu2024can}. Graph-Retrieval Augmented Generation (Graph-RAG) extends the Retrieval-Augmented Generation (RAG) framework \cite{lewis2020retrieval} by enriching large language models with structured knowledge retrieval. While traditional RAG retrieves unstructured text, Graph-RAG leverages GNNs to retrieve structured subgraphs encoded as triples, improving reasoning precision and minimizing redundancy \cite{guo2024knowledgenavigator, wen2023mindmap, lazaridou2022internet}.

Existing Graph-RAG benchmarks primarily evaluate basic graph reasoning tasks, such as shortest paths, node degree, and edge existence \cite{fatemi2023talk, wang2024can}. Although these benchmarks provide insights into foundational reasoning, they lack domain specificity. Recent work by He et al. \cite{he2024g} introduced benchmarks targeting advanced reasoning in general graph contexts, but domain-specific benchmarks for applications such as nutrition remain underdeveloped. By adapting the principles of Graph-RAG, our work introduces the first benchmark designed to tackle personalized health-aware reasoning, addressing this critical gap in the literature.

\section{Benchmark Details}
\label{appendix:Benchmark Details}

\subsection{Data Source Description}
\noindent\textbf{NHANES.} National Health and Nutrition Examination Survey (NHANES) is a publicly available dataset collected by the U.S. Centers for Disease Control and Prevention (CDC) to assess the health and nutritional status of the U.S. population through interviews, physical examinations, and laboratory tests. Data is released every two years and encompasses five main categories: Demographics, Dietary Data, Examination Data, Laboratory Data, and Questionnaire Data. These comprehensive datasets provide a wealth of information on health indicators, dietary behaviors, and medical conditions. 

\noindent\textbf{FNDDS and WWEIA.} The Food and Nutrient Database for Dietary Studies (FNDDS) is a comprehensive resource developed by the U.S. Department of Agriculture (USDA) to facilitate dietary intake analysis by providing detailed nutritional information for foods and beverages consumed in the United States. It serves as the backbone for analyzing dietary recall data collected through the What We Eat in America (WWEIA) program, which is a component of NHANES. WWEIA captures dietary intake data through 24-hour dietary recall interviews, linking reported food and beverage items to their corresponding nutrient profiles in FNDDS. Together, FNDDS and WWEIA enable researchers to study dietary patterns, nutrient intake, and their relationship to health outcomes, making them critical tools for advancing nutrition research and public health policy.

\subsection{Dietary Habit Processing Details}
Dietary habit data was sourced from various NHANES tables, including the Diet Behavior and Consumer Behavior datasets, which capture user-reported behaviors and preferences related to food choices, preparation methods, and consumption patterns. Traditional processing approaches proved insufficient for the complexity and diversity of these features. To address this, a thorough manual review was conducted by a team of four researchers. Key features indicative of dietary habits, such as awareness of healthy eating practices or frequency of consuming processed or frozen foods, were identified and categorized. Users were then grouped into high and low habit categories based on their responses, with the top 10\% and bottom 10\% assigned corresponding habit tags. For instance, users reporting the highest milk consumption were tagged with "drink lots of milk," while those with minimal consumption were labeled as "drink little or no milk." This process generated 54 distinct dietary habit tags, which were incorporated as nodes in the graph. These habit nodes provide critical insights into user behaviors, enabling a nuanced understanding of the relationship between dietary patterns and health outcomes.

\begin{table}[!t]
    \centering
    \setlength{\tabcolsep}{10pt} 
    \resizebox{\linewidth}{!}{
        \begin{tabular}{lccc}
            \toprule
            \textbf{Nutrients} & \textbf{Low Threshold} & \textbf{High Threshold} & \textbf{NRV} \\
            \midrule\midrule
            Calories (kcal)      & 40   & 225   & 2000 \\
            Carbohydrates (g)    & 55   & 75    & -    \\
            Protein (g)          & 10   & 15    & 50   \\
            Saturated Fat (g)    & 1.5  & 5     & 20   \\
            Cholesterol (mg)     & 20   & 40    & 300  \\
            Sugar (g)            & 5    & 22.5  & -    \\
            Dietary Fiber (g)    & 3    & 6     & -    \\
            \midrule\midrule
            Sodium (mg)          & 120  & 200   & 2000 \\
            Potassium (mg)       & 0    & 525   & 3500 \\
            Phosphorus (mg)      & 0    & 105   & 700  \\
            Iron (mg)            & 0    & 3.3   & 22   \\
            Calcium (mg)         & 0    & 150   & 1000 \\
            Folic Acid (µg)      & 0    & 60    & 400  \\
            Vitamin C (mg)       & 0    & 15    & 100  \\
            Vitamin D (µg)       & 0    & 2.25  & 15   \\
            Vitamin B12 (µg)     & 0    & 0.36  & 2.4  \\
            \bottomrule
        \end{tabular}
    }
    \caption{Nutrient Reference Values (NRV) and thresholds (per 100g of food) used based on the nutritional standards.}
    \vspace{-10pt}
    \label{tab:nutrient_thresholds}
\end{table}

\begin{table}[!t]
    \centering
    \setlength{\tabcolsep}{8pt} 
    \resizebox{\linewidth}{!}{
        \begin{tabular}{lcc}
            \toprule
            \textbf{Health Indicator} & \textbf{High Threshold} & \textbf{Low Threshold} \\
            \midrule\midrule
            BMI                         & 30           & 18.5         \\
            Waist Circumference (cm)    & 102 (88)     & -            \\
            Blood Pressure (mmHg)       & 140          & 90           \\
            Osteoporosis                & -            & -            \\
            \midrule
            Blood Urea Nitrogen (mmol/L) & 7.1         & -            \\
            Low-Density Lipoprotein (mmol/L) & 3.3      & -            \\
            Red Blood Cell (million cells/uL) & -       & 4            \\
            Glucose (mmol/L)            & 7            & -            \\
            Glycohemoglobin (\%)        & 6.5          & -            \\
            Hemoglobin (g/dL)           & -            & 13.2 (11.6)  \\
            \bottomrule
        \end{tabular}
    }
    \caption{Health Indicators with Corresponding High and Low Thresholds. Parentheses indicate sex-specific: male (female) thresholds where applicable.}
    \vspace{-10pt}
    \label{tab:health_indicators}
\end{table}

\subsection{Full Mappings of Nutrition Tags} 
In this section, we discuss the overall mapping relationship between health indicators and nutrition. In total, we involve nutrition tags for 16 different nutrients focusing on various health aspects, including 7 for macro-nutrients (calories, carbohydrates, protein, saturated fat, cholesterol, sugar, and dietary fiber) and 9 for micro-nutrients (sodium, potassium, phosphorus, iron, calcium, folic acid, and vitamin C, D, and B12) following the tagging scheme introduced in \cite{zhang2024mopi}. A detailed table of thresholds can be seen in Table-\ref{tab:nutrient_thresholds}. As discussed in the paper, these thresholds are derived from existing standards and legislation, from World Health Organization (WHO), Food Standards Agency (FSA)m EU Nutrition \& Health Claims Regulation \cite{EC2006} and the Codex Alimentarius Commission (CAC) \cite{FAO1985, FAO1997}. An even more detailed standards are listed in Appendix-\ref{appendix:Standards and Regulation}. Following the similar practice, we also extract the thresholds for health conditions, as shown in Table-\ref{tab:health_indicators}, Since we have the thresholds for both nutrition and health, we demonstrate the full mapping relationship can be seen in Table-\ref{tab:nutrient_tags_health_indicators}. Note that the special diet data can be retrieved from NHANES data, which directly indicates a user needs certain nutrients. 

However, as we emphasize in the paper, the interactions between nutrition and health are complex and multi-facet. To maintain scientific rigor and practical relevance, we focus on annotating four prevalent health statues, of which diet has been proved to be beneficial for intervention. Their mapping to nutrition tags can be seen in Table-\ref{tab:health_indicators_tags}. The definition of these major health statues are discussed in the next section.

\begin{table*}[!t]
    \centering
    \setlength{\tabcolsep}{6pt} 
    \resizebox{\textwidth}{!}{ 
        \begin{tabular}{llp{0.75\textwidth}}
            \toprule
            \textbf{Nutrient Category} & \textbf{Tag Name} & \textbf{Source Health Indicators} \\
            \midrule\midrule
            \multirow{9}{*}{Macro-nutrients} 
            & High Calories      & Low BMI; Low waist circumference; Weight gain/Muscle building diet \\
            & Low Calories       & High BMI; High waist circumference; Weight loss diet \\
            & Low Carb           & Low carbohydrate diet; High BMI; High waist circumference \\
            & High Protein       & Opioid misuse; Weight gain/Muscle building diet; High protein diet \\
            & Low Protein        & High blood urea nitrogen; Renal/Kidney diet \\
            & Low Saturated Fat  & High low-density lipoprotein; Low fat/Low cholesterol diet \\
            & Low Sugar          & Opioid misuse; Diabetic Diet; Low sugar Diet \\
            & Low Cholesterol    & High low-density lipoprotein; Low fat/Low cholesterol diet \\
            & High Fiber         & High low-density lipoprotein; Opioid misuse; Diabetic Diet \\
            \midrule
            \multirow{9}{*}{Micro-nutrients} 
            & Low Sodium         & High blood pressure; Renal/Kidney diet; Low salt diet \\
            & High Potassium     & High blood pressure \\
            & Low Phosphorus     & Renal/Kidney diet \\
            & High Iron          & Low red blood cell/Low hemoglobin \\
            & High Calcium       & Osteoporosis/brittle bones \\
            & High Folic Acid    & Low red blood cell count \\
            & High Vitamin C     & Low red blood cell/Low hemoglobin; Osteoporosis/brittle bones \\
            & High Vitamin D     & Osteoporosis/brittle bones \\
            & High Vitamin B12   & Low red blood cell count \\
            \bottomrule
        \end{tabular}
    }
    \caption{Nutrient Categories, Tag Names, and Associated Source Health Indicators. Nutrient categories are organized to consolidate related tags and their respective health indicators for clarity.}
    \vspace{-10pt}
    \label{tab:nutrient_tags_health_indicators}
\end{table*}

\begin{table}[!t]
    \centering
    \setlength{\tabcolsep}{8pt} 
    \resizebox{\linewidth}{!}{ 
        \begin{tabular}{ll}
            \toprule
            \textbf{Health Indicator} & \textbf{Associated Tags} \\
            \midrule
            Obesity                         & Low Calorie \\
            Opioid Misuse                   & High Protein; Low Sugar; Low Sodium \\
            Hypertension                    & Low Sodium \\
            Diabetes                        & Low Sugar; Low Carb \\
            Weight Loss/Low Calorie Diet    & Low Calorie \\
            Low Fat/Low Cholesterol Diet    & Low Cholesterol; Low Saturated Fat \\
            Low Salt/Low Sodium Diet        & Low Sodium \\
            Sugar-Free/Low Sugar Diet       & Low Sugar \\
            Diabetic Diet                   & Low Sugar; Low Carb \\
            Weight Gain/Muscle Building Diet & High Calorie; High Protein \\
            Low Carbohydrate Diet           & Low Carb \\
            High Protein Diet               & High Protein \\
            Renal/Kidney Diet               & Low Protein \\
            \bottomrule
        \end{tabular}
    }
    \caption{Health Indicators and Their Associated Nutritional Tags. Each indicator is linked to relevant tags reflecting dietary requirements.}
    \vspace{-10pt}
    \label{tab:health_indicators_tags}
\end{table}

\subsection{The Definition of Health Conditions} 
In the paper, we focus on annotating the four prevalent health statuses—obesity, hypertension, opioid misuse, and diabetes—that are directly influenced by dietary interventions. Among them, WHO and American Heart Association (AHA) provide clear and well-known definitions for obesity and hypertension. We mark a user obesity if the BMI is 30 or greater, and we mark a user hypertension if the average of 4 test of systolic pressure is 140 mm Hg or higher or diastolic pressure is 90 mm Hg or higher. This is classified as stage-2 hypertension and require medical control. For Diabetes, NHANES provides specific questionnaire for diabetic users, and we also mark a user diabetic if the user's Glucose (mmol/L) level is over 7.0 AND Glycohemoglobin (\%) is over 6.5. 

Opioid misuse, on the other hand, is a tricky health condition to be defined. However we argue this health condition is of vital importance, as the opioid crisis has been one of the most critical society concerns in the United States. Opioids are a category of drugs that include the illegal substance heroin, synthetic opioids such as fentanyl, and prescription painkillers like oxycodone \cite{NIDA-Opioid}. While primarily used for pain management, opioids can induce euphoria, making them prone to misuse \cite{dennett202opdiet,rigg2010motivations,rosenblum2008opioids}. For instance, in 2019, 10.1 million Americans reported opioid misuse, and in 2021, there were an estimated 108,000 drug overdose deaths in the United States, 90\% of which were linked to opioids \cite{CDC-OPmisuse, CDC-OPdeath}. In this work, we follow prior work \cite{zhang2024diet} to define misuse by the following criteria: (1) records of illicit opioid drug use, like heroin, within a year, or (2) records of prescription opioid medication use for over 90 days, which is a threshold commonly employed in the medical domain ~\cite{gu2022prevalence}. 

NHANES dataset provides illicit drug usage data, and we can track down the opioid prescription medicine usage data using the Multum Lexicon Therapeutic Classification Scheme, a 3-level nested category system that assigns a therapeutic classification to each drug and each ingredient of the drug. Category codes used to identify prescription opioid use were: Level 1: 57 = central nervous system agents; Level 2: 58 = Analgesics; Level 3: 60 = narcotic analgesics, or 191 = narcotic analgesics combinations (Detail in Appendix-\ref{appendix:Standards and Regulation}).

\begin{table*}[t]
\centering
\resizebox{\textwidth}{!}{
\begin{tabular}{cc|cccc|cccc|ccccc}
\toprule
\multirow{2}{*}{\textbf{Question Level}} & \multirow{2}{*}{\textbf{Method}} & \multicolumn{4}{c}{\textbf{a) Binary Classification (-B)}} & \multicolumn{4}{c}{\textbf{b) Multi-label Classification (-ML)}} & \multicolumn{5}{c}{\textbf{c) Text Generation (-TG)}} \\ 
\cmidrule{3-6} \cmidrule{7-10} \cmidrule{11-15}
 & & \textbf{Accuracy} & \textbf{Recall} & \textbf{Precision} & \textbf{F1} & \textbf{Accuracy} & \textbf{Recall} & \textbf{Precision} & \textbf{F1} & \textbf{ROUGE-1} & \textbf{ROUGE-2} & \textbf{ROUGE-L} & \textbf{BLEU} & \textbf{BERT} \\\midrule
\multirow{5}{*}{\textbf{Sparse}} 
& \textbf{Plain} & 0.6161 & 0.2413 & 0.8619 & 0.3770 & 0.2190 & 0.8958 & 0.2365 & 0.3666 & 0.5645 & 0.4999 & 0.5642 & 0.3092 & 0.9375 \\ 
& \textbf{KAPING} & 0.5329 & 0.0732 & 0.6268 & 0.1310 & 0.1951 & 0.8885 & 0.2194 & 0.3468 & 0.5374 & 0.4678 & 0.5370 & 0.2759 & 0.9346 \\ 
& \textbf{CoT-Zero} & 0.6049 & 0.2885 & 0.7255 & 0.4128 & 0.3633 & 0.7636 & 0.4265 & 0.5263 & 0.5593 & 0.5016 & 0.5589 & 0.3424 & 0.8871 \\ 
& \textbf{CoT-BAG} & 0.6060 & 0.2875 & 0.7307 & 0.4126 & \textbf{0.4204} & 0.7430 & \textbf{0.4724} & \textbf{0.5589} & 0.5479 & 0.4888 & 0.5474 & 0.3325 & 0.8849 \\ 
& \textbf{ToG} & \textbf{0.8483} & \textbf{0.6959} & \textbf{0.9844} & \textbf{0.8154} & 0.3227 & \textbf{0.9561} & 0.3168 & 0.4672 & \textbf{0.7216} & \textbf{0.6793} & \textbf{0.7215} & \textbf{0.4997} & \textbf{0.9582} \\ \midrule
\multirow{5}{*}{\textbf{Standard}} 
& \textbf{Plain} & 0.5903 & 0.2584 & 0.8871 & 0.4002 & 0.5651 & 0.9224 & 0.5665 & 0.6932 & 0.7746 & 0.7074 & 0.7344 & 0.5513 & 0.9656 \\ 
& \textbf{KAPING} & 0.4809 & 0.0480 & 0.6216 & 0.0891 & 0.4830 & 0.8954 & 0.5064 & 0.6391 & 0.7203 & 0.6368 & 0.6835 & 0.4748 & 0.9594 \\ 
& \textbf{CoT-Zero} & 0.6576 & 0.3528 & \textbf{1.0000} & 0.5216 & 0.5373 & 0.9963 & 0.5429 & 0.6948 & 0.7333 & 0.6446 & 0.7058 & 0.4940 & 0.9507 \\ 
& \textbf{CoT-BAG} & 0.5872 & 0.2197 & \textbf{1.0000} & 0.3603 & 0.5585 & \textbf{0.9984} & 0.5599 & 0.7084 & 0.5479 & 0.4888 & 0.5474 & 0.3325 & 0.8849 \\ 
& \textbf{ToG} & \textbf{0.8647} & \textbf{0.7443} & \textbf{1.0000} & \textbf{0.8534} & \textbf{0.8242} & 0.9238 & \textbf{0.8437} & \textbf{0.8745} & \textbf{0.8870} & \textbf{0.8292} & \textbf{0.8227} & \textbf{0.6959} & \textbf{0.9775} \\ \midrule
\multirow{5}{*}{\textbf{Complex}} 
& \textbf{Plain} & 0.6249 & 0.0424 & 0.3562 & 0.0758 & 0.6790 & 0.8679 & 0.7695 & 0.8108 & 0.7608 & 0.6814 & 0.7136 & 0.5102 & 0.9604 \\ 
& \textbf{KAPING} & 0.6302 & 0.0473 & 0.4143 & 0.0849 & 0.6549 & 0.8501 & 0.7522 & 0.7915 & 0.7446 & 0.6644 & 0.7032 & 0.4910 & 0.9587 \\ 
& \textbf{CoT-Zero} & 0.6639 & 0.0750 & 0.9787 & 0.1394 & 0.7466 & \textbf{0.9729} & 0.7693 & 0.8562 & 0.7474 & 0.6597 & 0.7107 & 0.5053 & 0.9475 \\ 
& \textbf{CoT-BAG} & 0.6621 & 0.0685 & \textbf{1.0000} & 0.1282 & \textbf{0.7533} & 0.9628 & 0.7783 & \textbf{0.8577} & 0.7468 & 0.6620 & 0.7076 & 0.5051 & 0.9470 \\ 
& \textbf{ToG} & \textbf{0.7219} & \textbf{0.2936} & 0.8295 & \textbf{0.4337} & 0.6871 & 0.7160 & \textbf{0.8952} & 0.7846 & \textbf{0.8177} & \textbf{0.7424} & \textbf{0.7651} & \textbf{0.5978} & \textbf{0.9692} \\ 
\bottomrule
\end{tabular}}
\vspace{-5pt}
\caption{Experimental results based on five baseline methods on the three tasks with the three question levels using the Llama-3.1-70B-instruct. The best performance of each group is bolded.}
\label{tab:experiment-results-llama}
\end{table*}

\begin{table*}[t]
\centering
\resizebox{\textwidth}{!}{
\begin{tabular}{cc|cccc|cccc|ccccc}
\toprule
\multirow{2}{*}{\textbf{Question Level}} & \multirow{2}{*}{\textbf{Method}} & \multicolumn{4}{c}{\textbf{a) Binary Classification (-B)}} & \multicolumn{4}{c}{\textbf{b) Multi-label Classification (-ML)}} & \multicolumn{5}{c}{\textbf{c) Text Generation (-TG)}} \\ 
\cmidrule{3-6} \cmidrule{7-10} \cmidrule{11-15}
 & & \textbf{Accuracy} & \textbf{Recall} & \textbf{Precision} & \textbf{F1} & \textbf{Accuracy} & \textbf{Recall} & \textbf{Precision} & \textbf{F1} & \textbf{ROUGE-1} & \textbf{ROUGE-2} & \textbf{ROUGE-L} & \textbf{BLEU} & \textbf{BERT} \\\midrule
\multirow{5}{*}{\textbf{Sparse}} 
& \textbf{Plain} & 0.5363 & 0.0384 & 0.9573 & 0.0739 & 0.1965 & 0.8102 & 0.2720 & 0.3770 & \textbf{0.4572} & \textbf{0.3806} & \textbf{0.4556} & \textbf{0.2137} & \textbf{0.9200} \\ 
& \textbf{KAPING} & 0.5370 & 0.0399 & \textbf{0.9588} & 0.0766 & 0.1960 & 0.8120 & 0.2713 & 0.3769 & 0.4565 & 0.3798 & 0.4548 & 0.2135 & 0.9199 \\ 
& \textbf{CoT-Zero} & 0.5324 & 0.0301 & 0.9535 & 0.0583 & 0.2535 & 0.8273 & \textbf{0.3934} & 0.4664 & 0.4350 & 0.3575 & 0.4334 & 0.1992 & 0.8728 \\ 
& \textbf{CoT-BAG} & 0.5885 & 0.2983 & 0.6607 & 0.4110 & \textbf{0.2698} & \textbf{0.8720} & 0.3523 & \textbf{0.4693} & 0.4498 & 0.3767 & 0.4485 & 0.2116 & 0.8777 \\ 
& \textbf{ToG} & \textbf{0.6336} & \textbf{0.4025} & 0.7109 & \textbf{0.5140} & 0.2100 & 0.7045 & 0.2493 & 0.3563 & 0.4480 & 0.3441 & 0.4432 & 0.1940 & 0.9074 \\ 
\midrule
\multirow{5}{*}{\textbf{Standard}} 
& \textbf{Plain} & 0.5268 & 0.1054 & \textbf{1.0000} & 0.1907 & 0.4599 & 0.8212 & 0.5386 & 0.6216 & 0.6260 & 0.5178 & 0.6067 & 0.3607 & 0.9380 \\ 
& \textbf{KAPING} & 0.5245 & 0.1007 & \textbf{1.0000} & 0.1830 & 0.4606 & 0.8214 & 0.5396 & 0.6228 & 0.6272 & 0.5192 & \textbf{0.6076} & \textbf{0.3623} & \textbf{0.9387} \\ 
& \textbf{CoT-Zero} & 0.4917 & 0.0391 & \textbf{1.0000} & 0.0753 & 0.5280 & 0.8426 & 0.6216 & 0.6881 & 0.5854 & \textbf{0.5708} & 0.4747 & 0.3213 & 0.9120 \\ 
& \textbf{CoT-BAG} & 0.5953 & 0.3100 & 0.8049 & 0.4476 & \textbf{0.5654} & \textbf{0.8577} & \textbf{0.6222} & \textbf{0.7073} & 0.6147 & 0.5128 & 0.5968 & 0.3504 & 0.9184 \\ 
& \textbf{ToG} & \textbf{0.8385} & \textbf{0.7630} & 0.9178 & \textbf{0.8333} & 0.5151 & 0.7613 & 0.5774 & 0.6378 & \textbf{0.6302} & 0.5061 & 0.5985 & 0.3526 & 0.9284 \\ 
\midrule
\multirow{5}{*}{\textbf{Complex}} 
& \textbf{Plain} & 0.6627 & 0.0799 & 0.8909 & 0.1467 & 0.5991 & \textbf{0.7924} & 0.7511 & 0.7482 & 0.6636 & \textbf{0.5725} & 0.6432 & \textbf{0.3953} & \textbf{0.9402} \\ 
& \textbf{KAPING} & 0.6645 & 0.0865 & 0.8833 & 0.1575 & 0.5998 & 0.7884 & 0.7518 & 0.7458 & \textbf{0.6637} & 0.5713 & \textbf{0.6452} & 0.3934 & 0.9400 \\ 
& \textbf{CoT-Zero} & 0.6467 & 0.0277 & \textbf{0.9444} & 0.0539 & \textbf{0.6352} & 0.7831 & \textbf{0.8071} & \textbf{0.7761} & 0.6300 & 0.5339 & 0.6149 & 0.3574 & 0.9184 \\ 
& \textbf{CoT-BAG} & 0.6556 & 0.2186 & 0.5654 & 0.3153 & 0.6295 & 0.7686 & 0.7996 & 0.7712 & 0.6506 & 0.5619 & 0.6321 & 0.3829 & 0.9223 \\ 
& \textbf{ToG} & \textbf{0.7710} & \textbf{0.7732} & 0.6565 & \textbf{0.7101} & 0.5224 & 0.6157 & 0.7529 & 0.6408 & 0.6296 & 0.5114 & 0.5981 & 0.3500 & 0.9267 \\ 
\bottomrule
\end{tabular}}
\vspace{-5pt}
\caption{Experimental results based on five baseline methods on the three tasks with the three question levels using the GPT-3.5-turbo. The best performance of each group is bolded.}
\vspace{-15pt}
\label{tab:experiment-results-3.5}
\end{table*}

\subsection{Definitions of Ground Truth}
In this section, we outline how ground truths are determined for each task. For the multi-label classification task, the process is straightforward. As discussed earlier, nutrition tags are created and linked to users’ health conditions based on predefined standards. The ground truths for this task are simply the lists of nutrition tags relevant to each user’s health profile.

For the binary classification task, we use the relationship between the user’s condition and the food’s nutrition tags. A "Yes" label is assigned if the relationship is a "match," and "No" is assigned if the relationship is a "contradict." In the case of complex question settings, where multiple "match" and "contradict" links exist, we calculate the count of each. A question is marked as "Yes" if the number of "match" links exceeds the number of "contradict" links.

For the text generation task, we generate reference texts using a combined approach. First, the overall healthiness of the food is determined using the binary classification result ("Yes" or "No"). This is followed by a natural language explanation that lists the relevant nutrition tags. For example, a reference text might read: "Yes, because the food is low in calories and high in protein." This method ensures that the reference text provides a clear and natural explanation for the decision.

\section{Implementation Details}
\label{appendix:Implementation Details}

In this section, we discuss the implementation details of the baseline models. Specially how we set the hyper-parameters and how we make adaption to our task. All codes all provided in the codebase mentioned in the abstract. 

\textbf{Plain} refers to a naive GraphRAG pipeline. Unlike approaches that directly input natural language text or tabular data, we transform the user and food information from the knowledge graph structure into multiple triples, each consisting of an entity, a relationship, and another entity, then concatenate them before feeding into the LLMs.

\textbf{KAPING} answers questions based on a subgraph composed of the entities mentioned in the query and their neighboring nodes. Following the methodology described in the original paper, we first extract the entities present in the query—specifically the user and food—from the provided knowledge graph. Then, we include their respective neighboring nodes to construct a subgraph via retrieval. This subgraph is subsequently transformed into triples and concatenated before feeding into the LLMs. Note that in the original implementation, the authors also used top-k filtering to prune the retrieval results. However, since we don't have any other entities in the question, this pruning based on embedding similarities with the question doesn't generate any reasonable results. We skip this step in our implementation.

\textbf{CoT-Zero} is a two-stage prompting stategy. In the first stage, "Let's think step by step" is appended after the question to guide the model towards producing a reasoning path. In the second stage, the reasoning path is fed to the model to extract the final answer. However, our initial experiments showed that we can combine these two steps, by having both "Let's think step by step" and final output requirements in one prompt, while still achieving the same performance. This allows us to save computational and API resources, avoiding potential inconsistencies and information loss that arise when feeding the reasoning output into a second step. This is because with the one-step approach, the model can make a final decision based on both the original graph, and its own reasoning path, whereas in the second-step approach, the original graph is not available to the model.

\textbf{CoT-BAG} is designed to improve the graph reasoning capabilities of LLMs by first encouraing the model to "build" an implicit graph representation of the problem, and then using chain-of-thought reasoning to solve it. For this approach, a single prompt is sufficient to guide the model through both the graph construction and reasoning, by combining both "Let's construct a graph from the given nodes and edges" and "Let's think step by step to arrive at the final answer". Adapting CoT-BaG to our benchmark requires creating a textual description of the graph triples, in the following format: "The graph contains an edge between node [source] and node [target] with attribute [relationship], an edge between..." to include in the input prompt, alongside the question, and output requirements.

\textbf{ToG} introduces a strategy that iteratively searches and prunes reasoning paths on a knowledge graph starting from entities mentioned in the query to identify suitable paths. However, the open-source ToG codebase is implemented based on Wikidata and Freebase databases, making it incompatible with private datasets. To evaluate ToG on our benchmark, we reimplemented it following the original methodology. Furthermore, we adapted ToG to better suit the characteristics of our benchmark with the following adjustments: 1). Adjusting the width parameter to 5: ToG’s original width parameter is set to 3, which retains three reasoning paths during pruning. However, answering questions in our benchmark sometimes requires more than three reasoning paths. By setting the width parameter to 5, ToG preserves five reasoning paths at each pruning step and generates answers based on these paths. 2). Delaying pruning until the second iteration: In ToG’s first iteration, the information gathered is often insufficient to evaluate the importance of each reasoning path. Pruning too early risks discarding paths that may be critical for answering the query. Delaying pruning allows ToG to collect more comprehensive information before making pruning decisions. These modifications ensure that ToG is better aligned with the requirements and complexities of our benchmark, enabling more effective performance evaluation.

\section{Additional Experiments}
\label{appendix:Additional Experiments}

\begin{table*}[!t]
    \centering
    \setlength{\tabcolsep}{8pt} 
    \resizebox{\textwidth}{!}{ 
        \begin{tabular}{lcccc}
            \toprule
            \textbf{Diet Type} & \textbf{Obesity} & \textbf{Hypertension} & \textbf{Opioid Misuse} & \textbf{Diabetes} \\
            \midrule
            Weight Loss/Low Calorie Diet      & 2,253 & 647 & 222 & 267 \\
            Low Fat/Low Cholesterol Diet      & 448   & 247 & 76  & 116 \\
            Low Salt/Low Sodium Diet          & 442   & 350 & 86  & 115 \\
            Sugar-Free/Low Sugar Diet         & 170   & 89  & 20  & 78  \\
            Diabetic Diet                     & 692   & 432 & 126 & 647 \\
            Weight Gain/Muscle Building Diet  & 3     & 20  & 12  & 1   \\
            Low Carbohydrate Diet             & 244   & 69  & 25  & 57  \\
            High Protein Diet                 & 47    & 12  & 9   & 8   \\
            Renal/Kidney Diet                 & 25    & 24  & 13  & 7   \\
            \bottomrule
        \end{tabular}
    }
    \caption{Adoption of Diet Types Across Health Conditions. Each entry represents the number of users with a specific condition following a corresponding diet type.}
    \label{tab:condition_diet_distribution}
\end{table*}

\begin{table}[!t]
    \centering
    \setlength{\tabcolsep}{8pt} 
    \resizebox{\linewidth}{!}{ 
        \begin{tabular}{lc}
            \toprule
            \textbf{Status} & \textbf{\# Users} \\
            \midrule
            Weight Loss/Low Calorie Diet   & 4,693 \\
            Low Fat/Low Cholesterol Diet   & 1,196 \\
            Low Salt/Low Sodium Diet       & 1,037 \\
            Sugar-Free/Low Sugar Diet      & 417   \\
            Diabetic Diet                  & 1,403 \\
            Weight Gain/Muscle Building Diet & 274 \\
            Low Carbohydrate Diet          & 489   \\
            High Protein Diet              & 146   \\
            Renal/Kidney Diet              & 59    \\
            Obesity                        & 18,271 \\
            Hypertension                   & 10,257 \\
            Opioid Misuse                  & 2,822 \\
            Diabetes                       & 3,837 \\
            \bottomrule
        \end{tabular}
    }
    \caption{Distribution of Users Across Health Conditions and Special Diets.}
    \vspace{-10pt}
    \label{tab:user_status_distribution}
\end{table}

To further demonstrate the performance of different LLM backbones on our benchmark, we conducted additional tests using Llama-3.1-70b-Instruct and GPT-3.5-Turbo as backbones for various baselines. As shown in Table-\ref{tab:experiment-results-llama} and Table-\ref{tab:experiment-results-3.5}, the performance trends of Llama-3.1-70b-Instruct align closely with those of GPT-4o-mini, although Llama-3.1-70b-Instruct generally yields better results. This is consistent with its stronger reasoning capabilities.

Additionally, ToG exhibited a noticeable performance degradation when GPT-3.5-Turbo was used as the backbone, particularly when addressing standard and complex questions. This decline is primarily due to GPT-3.5-Turbo’s relatively weaker reasoning abilities, which often lead to the retrieval of suboptimal information. Such information provides minimal support—or even introduces negative impacts—on subsequent answer generation. These two sets of experiments highlight the stringent reasoning requirements imposed by our benchmark on the tested models.

\section{Additional Statistics}
\label{appendix:Additional Statistics}

In addition to the basic statistics provided above, we also provide an in detailed benchmark discussing the user distribution on health conditions and the overlap between the four major conditions and the special diets.  

Spanning from 2003 to 2020, the latest available NHANES data includes a total of 95,872 unique users. Table-\ref{tab:user_status_distribution} illustrates the distribution of health conditions across this population, highlighting the significant prevalence of obesity (18,271 users) and hypertension (10,257 users). These numbers emphasize the widespread impact of these conditions on public health and underscore the urgent need for dietary interventions. However, the stark contrast between the prevalence of these conditions and the adoption of relevant dietary interventions—such as low-calorie diets (4,693 users) or low-sodium diets (1,037 users)—reveals a significant gap. While conditions like obesity and hypertension demand immediate dietary action, far fewer individuals engage in corresponding interventions. This disparity highlights the critical need for personalized dietary reasoning to encourage healthier eating habits tailored to individual health conditions.

A similar trend emerges in Table-\ref{tab:condition_diet_distribution}, which examines the alignment between specific health conditions and diet types. While there is some adoption of relevant dietary actions, such as weight loss diets (2,253 for obesity, 647 for hypertension) and low-sodium diets (442 for obesity, 350 for hypertension), these numbers remain disproportionately low relative to the overall prevalence of these conditions. The gap is even more pronounced for diabetes, where fewer than half of diagnosed individuals (647 users) follow diabetic diets out of 3,837 diagnosed users. Specialized interventions, such as renal/kidney or muscle-building diets, see minimal adoption across all conditions, suggesting a lack of accessibility or awareness for these targeted approaches. These patterns reinforce the need for tailored, actionable dietary recommendations to address the divide between health condition prevalence and effective dietary responses, ensuring broader access to appropriate and impactful interventions.

\begin{figure*}[t!]
  \centering
  \includegraphics[width=\textwidth]{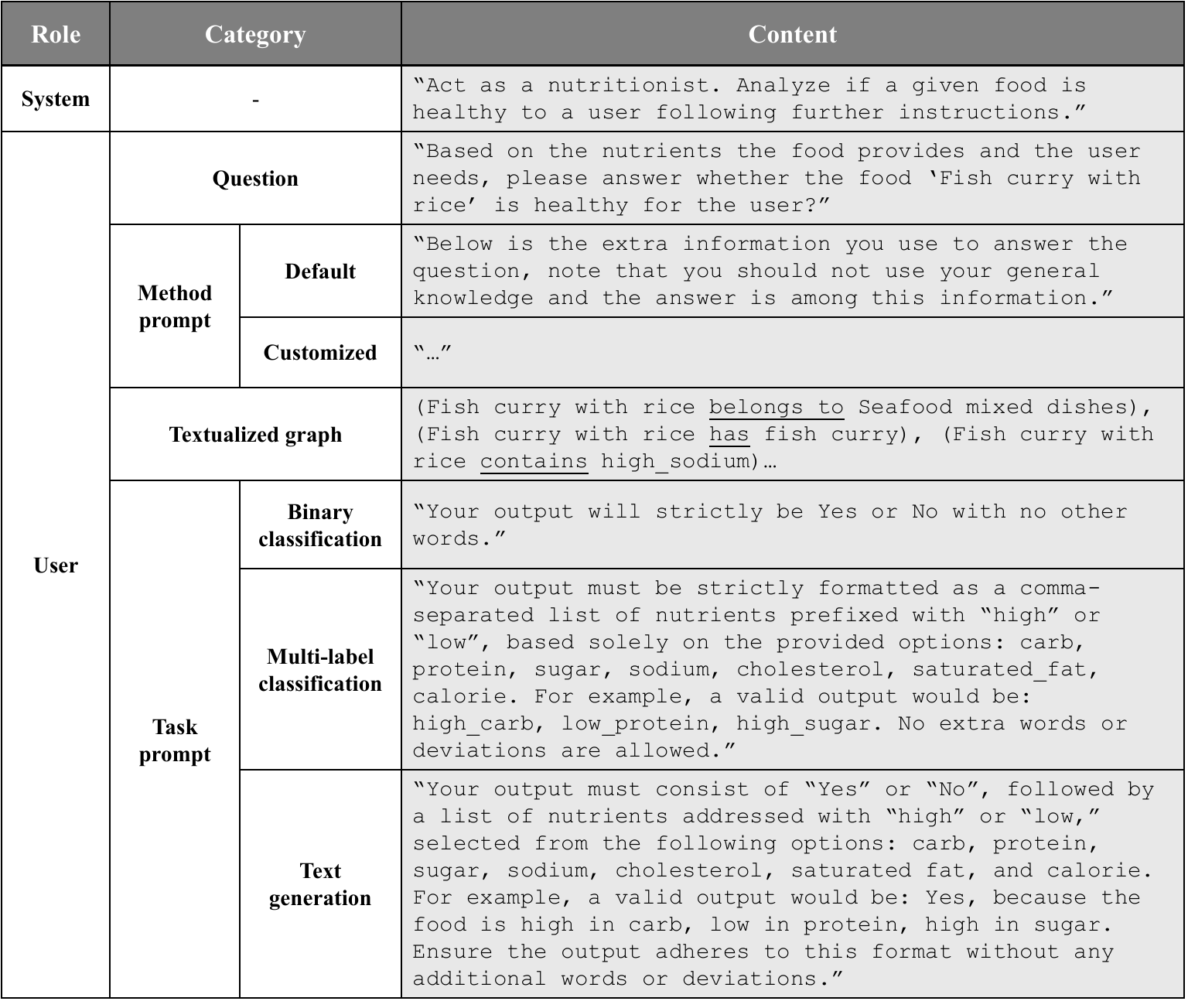}
  \vspace{-15pt}
  \caption{The paradigm of prompt for final output.}
  \vspace{-15pt}
  \label{fig:prompt_all}
\end{figure*}

\begin{figure}[t!]
  \centering
  \includegraphics[width=\columnwidth]{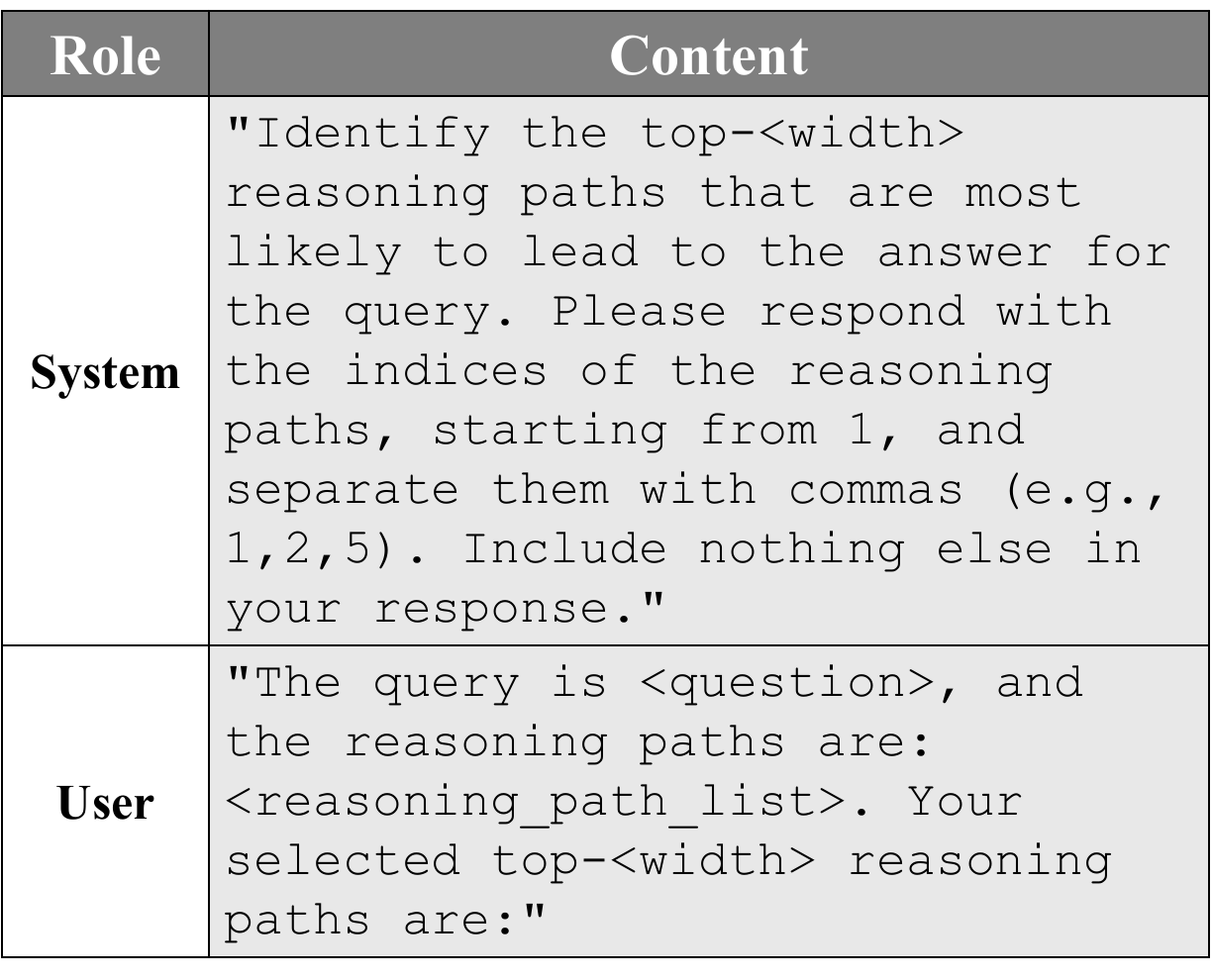}
  \vspace{-15pt}
  \caption{The prompt used in ToG.}
  \vspace{-15pt}
  \label{fig:prompt_tog}
\end{figure}

\section{Prompt Design}
\label{appendix:Prompt Design}

In this section, we will demonstrate our carefully designed prompts for the three task settings and selected baselines. The principle of our prompt design is to let LLMs become familiar with nutritional domain knowledge while avoiding providing explicit guidance.

When querying LLMs for the final output, the paradigm of our prompt is shown as Figure-\ref{fig:prompt_all}. The system prompt is fixed while the user prompt consists of four flexible parts: \textit{question}, \textit{method prompt}, \textit{textualized graph}, and \textit{task prompt}. The \textit{question} and \textit{task prompt} will be automatically adjusted according to the experiment settings. The \textit{method prompt} can be customized to the methods proposed by the benchmark users, e.g., adding "Let's think step by step." for CoT-Zero and adding "Let’s construct a graph from the given nodes and edges" for CoT-BAG. We encourage benchmark users to further explore the potential of method prompts. The \textit{textualized graph} is by default generated by concatenating the triplets in the retrieved knowledge graph. Benchmark users can also customize their own textualization method.

Additionally, the prompt we used to prune the relations and entities when testing ToG is shown in Figure-\ref{fig:prompt_tog}. 

\section{Case Study}
\label{appendix:Case Study}

We present 7 case studies across 3 Tasks (Binary Classification, Multi-label Classification, Text Generation), 3 Question Levels (Sparse, Standard, Complex) and 5 Baselines (Plain, KAPING, CoT-Zero, CoT-BaG, ToG). This section provides insights into how the prompts are structured across different baselines and the reasoning path behind the LLM's final answer, as detailed in Tables \ref{tab:case_1}-\ref{tab:case_7}. The case studies provide critical insights into the strengths and limitations of each baseline, while emphasizing the challenges posed by personalized dietary reasoning, highlighting our benchmark’s role in advancing the development of robust, domain-specific AI models for personalized health-aware nutrition reasoning. 


\begin{table*}[t]
\centering
\small
\renewcommand{\arraystretch}{1.5}
\begin{tabular}{lp{12cm}}
\toprule
\textbf{Configurations} & Task: Binary Classification \\
                        & Question: Complex \\
                        & Model: GPT-4o-mini \\
                        & Method: Plain \\ \midrule
\textbf{Prompt} & 
\textit{[Question]} Based on the nutrients the food provides and the user needs, please answer whether the food \textbf{"Black beans and brown rice"} is healthy for the user? Please answer with yes or no.

\vspace{5pt}

\textit{[Method prompt]} Below are the extra information you use to answer the question, note that you should not use your general knowledge and the answer is among this information.

\vspace{5pt}

\textit{[Textualized graph]} 
(Black beans and brown rice belongs to Bean, pea, legume dishes), (Black beans and brown rice has Black beans, from dried, fat added), (Black beans and brown rice has Rice, brown, cooked, no added fat), (Black beans and brown rice belongs to low\_carb), (Black beans and brown rice belongs to low\_sugar), (Black beans and brown rice belongs to high\_sodium), (Black beans and brown rice belongs to low\_protein), (Black beans and brown rice belongs to low\_cholesterol), (Black beans and brown rice belongs to low\_saturated\_fat), (user has Eats lots of fish), (user has Adds little to no salt at table), (user has Eats little to no fast food), (user has Eats few to no ready to eat meals), (user has Takes more supplements), (user has Uses lots of salt in preparation), (user has Ate more food than usual), (user has Ate less food than usual), (user has Eats gluten free diet), (user has Eats high protein diet), (user has diabetes), (user has Low carbohydrate diet), (user has High protein diet), (diabetes match low\_sugar), (diabetes match low\_carb), (Low carbohydrate diet match low\_carb), (High protein diet contradict low\_protein)

\vspace{5pt}

\textit{[Note prompt]} Important Note: Your output will strictly be Yes or No with no other words. \\ \midrule
\textbf{True Answer} & No \\ \midrule
\textbf{LLM Answer} & \textcolor{ForestGreen}{No} \\ \midrule
\textbf{LLM Reasoning} & 
While black beans and brown rice provide some beneficial nutrients, they are low in protein, which contradicts the user's high protein diet requirement. \textcolor{red}{Additionally, the dish is high in sodium, which may not be suitable for someone managing diabetes and following a low carbohydrate and low sugar diet}. Therefore, it does not align well with the user's nutritional needs. \\ \midrule
\textbf{Observation} & 
Factual hallucination: Although the model is able to correctly determine that the food is unhealthy to the user because it is low in protein compared to the user’s High protein diet requirement, upon inspecting its reasonings, it is found that the model mistakenly deduces that high\_sodium food tag is unsuitable for a low carb and low sugar diet, which is not correct. \\
\bottomrule
\end{tabular}
\caption{Case Study 1}
\label{tab:case_1}
\end{table*}


\begin{table*}[t]
\centering
\small
\renewcommand{\arraystretch}{1.5}
\begin{tabular}{lp{12cm}}
\toprule
\textbf{Configurations} & Task: Multi-label Classification \\
                        & Question: Standard \\
                        & Model: Llama-3.1-70b-instruct \\
                        & Method: KAPING \\ \midrule
\textbf{Prompt} & 
\textit{[Question]} Based on the nutrients the food provides and the user needs, please answer what nutrient tags are used to determine whether the food \textbf{"Nachos, cheese only"} is healthy or unhealthy for the user? 

\vspace{5pt}

\textit{[Method prompt]} Below are the extra information you use to answer the question, note that you should not use your general knowledge and the answer is among this information.

\vspace{5pt}

\textit{[Textualized graph]} 
(Nachos, cheese only belongs to Nachos), (Nachos, cheese only has Cheese sauce), (Nachos, cheese only has Snacks, tortilla chips, plain, white corn, salted), (Nachos, cheese only has Fast foods, nachos, with cheese), (Nachos, cheese only belongs to low\_carb), (Nachos, cheese only belongs to low\_sugar), (Nachos, cheese only belongs to high\_sodium), (Nachos, cheese only belongs to high\_calorie), (Nachos, cheese only belongs to low\_protein), (Nachos, cheese only belongs to low\_cholesterol), (user has Drinks little or no milk), (user has Eats little or no shellfish), (user has Adds little to no salt at table), (user has Drinks Alcohol more than average), (user has Eats many ready to eat meals), (user has Drinks lots of water), (user has Ate more food than usual), (user has Ate less food than usual), (user has Eats gluten free diet), (user has obesity), (user has High protein diet), (obesity contradict high\_calorie), (High protein diet contradict low\_protein)

\vspace{5pt}

\textit{[Note prompt]} Important Note: Your output must be strictly, with no extra words, separated by comma, a list of nutrients with high or low before the nutrients among these options: carb, protein, sugar, sodium, cholesterol, saturated\_fat, calorie. For example, the output is: high\_carb, low\_protein, high\_sugar. You should only include the nutrient tags that connect the food with the user. \\ \midrule
\textbf{True Answer} & high\_calorie, low\_protein \\ \midrule
\textbf{LLM Answer} & \textcolor{ForestGreen}{high\_calorie}, \textcolor{red}{high\_sodium}, \textcolor{ForestGreen}{low\_protein} \\ \midrule
\textbf{LLM Reasoning} & 
It is stated that "Nachos, cheese only belongs to high\_calorie" and "high\_sodium." Additionally, it is mentioned that "Nachos, cheese only belongs to low\_protein."

Given the user's condition of obesity and High protein diet requirements, consuming high-calorie food would contradict their health goal of obesity. Furthermore, a high-protein diet requirement contradicts the fact that “Nachos, cheese only belongs to low\_protein”. \textcolor{red}{User’s habit of adding little to no salt at table contradicts the fact that “Nachos, cheese only belong to high\_sodium”.} \\ \midrule
\textbf{Notes} & The model includes information about the user’s habit of adding little to no salt at the table and concludes that high\_sodium is a food tag that is important in determining the food’s healthiness, which is irrelevant information in this case because it is supposed to focus on the user’s health and diet only - \textit{Factual Hallucination}. \\
\bottomrule
\end{tabular}
\caption{Case Study 2}
\label{tab:case_2}
\end{table*}


\begin{table*}[t]
\centering
\small
\renewcommand{\arraystretch}{1.5}
\begin{tabular}{lp{12cm}}
\toprule
\textbf{Configurations} & Task: Text Generation \\
                        & Question: Complex \\
                        & Model: GPT-4o-mini \\
                        & Method: CoT-Zero \\ \midrule
\textbf{Prompt} & 
\textit{[Question]} Based on the nutrients the food provides and the user needs, please answer whether the food \textbf{"Turkey with gravy"} is healthy for the user? Please answer with a short sentence explaining why.

\vspace{5pt}

\textit{[Method prompt]} Below are the extra information you use to answer the question, note that you should not use your general knowledge and the answer is among this information. Let's think step by step to determine the healthiness of the food, by extracting the nutritional properties of the food from the given graph, then comparing them to the nutrition requirements of the health status, dietary need and habits of the user. A food is unhealthy only if it has certain properties that are unsuitable to the user's health and diet. Do not be too strict with your criteria, only focus on a few main nutritional tags that strongly indicate its healthiness or unhealthiness to the particular diet or health status the user has. Some nutritional tags might not be as important in determining healthiness.

\vspace{5pt}

\textit{[Textualized graph]} 
(Turkey with gravy belongs to Poultry mixed dishes), (Turkey with gravy has Turkey, whole, meat only, cooked, roasted), (Turkey with gravy has Salt, table, iodized), (Turkey with gravy has Gravy, chicken, canned or bottled, ready-to-serve), (Turkey with gravy belongs to low\_carb), (Turkey with gravy belongs to low\_sugar), (Turkey with gravy belongs to high\_sodium), (Turkey with gravy belongs to high\_protein), (Turkey with gravy belongs to high\_cholesterol), (Turkey with gravy belongs to low\_saturated\_fat), (user has Eats little or no shellfish), (user has Drinks Alcohol less than average), (user has Eats little to no frozen food), (user has Eats few to no meals outside home), (user has Eats few to no ready to eat meals), (user has Takes few or no supplements), (user has Uses little to no salt in preparation), (user has Ate more food than usual), (user has Ate less food than usual), (user has Eats weight loss diet), (user has Eats low fat diet), (user has Eats high fiber diet), (user has opioid\_misuse), (user has diabetes), (user has Low fat/Low cholesterol diet), (user has Low salt/Low sodium diet), (user has Diabetic diet), (opioid\_misuse match high\_protein), (opioid\_misuse match low\_sugar), (opioid\_misuse contradict high\_sodium), (diabetes match low\_sugar), (diabetes match low\_carb), (Low fat/Low cholesterol diet contradict high\_cholesterol), (Low fat/Low cholesterol diet match low\_saturated\_fat), (Low salt/Low sodium diet contradict high\_sodium), (Diabetic diet match low\_sugar), (Diabetic diet match low\_carb)

\vspace{5pt}

\textit{[Note prompt]} Important Note: Your output must be a Yes or No followed by strictly a list of nutrients with high or low as prefix among these options: carb, protein, sugar, sodium, cholesterol, saturated fat, calorie. For example, the output is: Yes, because the food is high in carb, low in protein, high in sugar. \\ \midrule
\textbf{True Answer} & Yes, because the food is low in carb, low in sugar, low in saturated\_fat, high in protein. \\ \midrule
\textbf{LLM Answer} & \textcolor{red}{No}, because the food is \textcolor{red}{high in sodium, high in cholesterol.} \\ \midrule
\textbf{LLM Reasoning} & 
Turkey with gravy is high in sodium, which contradicts the user's low salt/low sodium diet. It is also high in cholesterol, which contradicts the user’s low fat/low cholesterol diet. \\ \midrule
\textbf{Notes} & The model fails to recognize there are far more good properties about this food item than bad ones. \\
\bottomrule
\end{tabular}
\caption{Case Study 3}
\label{tab:case_3}
\end{table*}


\begin{table*}[t]
\centering
\small
\renewcommand{\arraystretch}{1.5}
\begin{tabular}{lp{12cm}}
\toprule
\textbf{Configurations} & Task: Multi-label Classification \\
                        & Question: Sparse \\
                        & Model: GPT-3.5-turbo \\
                        & Method: CoT-BaG \\ \midrule
\textbf{Prompt} & 
\textit{[Question]} Based on the nutrients the food provides and the user needs, please answer whether the food \textbf{"Sesame chicken"} is healthy for the user? Please answer with a short sentence explaining why.

\vspace{5pt}

\textit{[Method prompt]} Below are the extra information you use to answer the question, note that you should not use your general knowledge and the answer is among this information. You will be given the textual description of a directed graph. Let's first construct a graph with the given nodes and edges. Then determine the healthiness of the food by traversing the graph and determining the nutritional properties of the food, then compare them to the health status, dietary need and habits of the user. Do not be too strict with your criteria, only focus on a few main nutritional tags that strongly indicate its healthiness or unhealthiness to the particular diet or health status the user has. Some nutritional tags might not be as important in determining healthiness.

\vspace{5pt}

\textit{[Textualized graph]} 
Here is the description of the graph:
This is the list of edges: an edge between node "Sesame chicken" and "Stir-fry and soy-based sauce mixtures" with attribute "belongs to", an edge between node "Sesame chicken" and "Restaurant, Chinese, sesame chicken" with attribute "has", an edge between node "Sesame chicken" and "low\_carb" with attribute "belongs to", an edge between node "Sesame chicken" and "high\_sodium" with attribute "belongs to", an edge between node "Sesame chicken" and "high\_calorie" with attribute "belongs to", an edge between node "Sesame chicken" and "high\_cholesterol" with attribute "belongs to", an edge between node "user" and "Eats little or no shellfish" with attribute "has", an edge between node "user" and "Eats little or no fish" with attribute "has", an edge between node "user" and "Adds little to no salt at table" with attribute "has", an edge between node "user" and "Heavy cigarette smoker" with attribute "has", an edge between node "user" and "Drinks Alcohol more than average" with attribute "has", an edge between node "user" and "Eats lots of frozen food" with attribute "has", an edge between node "user" and "Eats few to no ready to eat meals" with attribute "has", an edge between node "user" and "Takes few or no supplements" with attribute "has", an edge between node "user" and "Drinks lots of water" with attribute "has", an edge between node "user" and "Eats organic food less often" with attribute "has", an edge between node "user" and "Claims to have a poor diet" with attribute "has", an edge between node "user" and "Ate more food than usual" with attribute "has", an edge between node "user" and "obesity" with attribute "has", an edge between node "user" and "hypertension" with attribute "has", an edge between node "obesity" and "high\_calorie" with attribute "contradict", an edge between node "hypertension" and "high\_sodium" with attribute "contradict".

\vspace{5pt}

\textit{[Note prompt]} Important Note: Your output must be a Yes or No followed by strictly a list of nutrients with high or low as prefix among these options: carb, protein, sugar, sodium, cholesterol, saturated fat, calorie. For example, the output is: Yes, because the food is high in carb, low in protein, high in sugar. \\ \midrule
\textbf{True Answer} & No, because the food is high in sodium, high in calorie. \\ \midrule
\textbf{LLM Answer} & No, because the food is \textcolor{ForestGreen}{high in sodium, high in calorie}, \textcolor{red}{high in cholesterol}. \\ \midrule
\textbf{LLM Reasoning} & 
Sesame chicken typically contains a lot of sodium due to the soy-based sauce, high calories from frying, and cholesterol from the meat and sauce. These nutritional properties may not align well with the user's dietary needs, especially if they have hypertension or obesity. \\ \midrule
\textbf{Notes} & Although the model is able to identify two important food tags that contribute to the food’s unhealthiness, upon inspecting its reasonings, it is unclear whether the model deduces this answer from the graph information, or from its general knowledge of the food’s ingredients. \\
\bottomrule
\end{tabular}
\caption{Case Study 4}
\label{tab:case_4}
\end{table*}


\begin{table*}[t]
\centering
\small
\renewcommand{\arraystretch}{1.5}
\begin{tabular}{lp{12cm}}
\toprule
\textbf{Configurations} & Task: Text Generation \\
                        & Question: Standard \\
                        & Model: GPT-4o-mini \\
                        & Method: ToG \\ \midrule
\textbf{Prompt} & 
\textit{[Question]} Based on the nutrients the food provides and the user needs, please answer whether the food \textbf{"Vegetable rice soup, canned, prepared with water or ready-to-serve"} is healthy for the user? Please answer with a short sentence explaining why.

\vspace{5pt}

\textit{[Method prompt]} Below are the extra information you use to answer the question, note that you should not use your general knowledge and the answer is among this information. You will be given the textual description of a directed graph.

\vspace{5pt}

\textit{[Textualized graph]} \textit{Note: Bolded triples are those that appear in ToG’s retrieved subgraph.}

\vspace{3pt}

(Vegetable rice soup, canned, prepared with water or ready-to-serve belongs to Soups), (Vegetable rice soup, canned, prepared with water or ready-to-serve has Soup, minestrone, canned, chunky, ready-to-serve), \textbf{(Vegetable rice soup, canned, prepared with water or ready-to-serve belongs to low\_carb)}, (Vegetable rice soup, canned, prepared with water or ready-to-serve belongs to low\_sugar), (Vegetable rice soup, canned, prepared with water or ready-to-serve belongs to high\_sodium), (Vegetable rice soup, canned, prepared with water or ready-to-serve belongs to low\_protein), \textbf{(Vegetable rice soup, canned, prepared with water or ready-to-serve belongs to low\_cholesterol)}, \textbf{(Vegetable rice soup, canned, prepared with water or ready-to-serve belongs to low\_saturated\_fat)}, (user has Drinks lots of milk), (user has Eats little or no shellfish), (user has Eats little or no fish), (user has Adds little to no salt at table), (user has Eats few to no meals outside home), (user has Eats few to no ready to eat meals), (user has Takes more supplements), (user has Claims to have a good diet), (user has Ate more food than usual), (user has Ate less food than usual), (user has Eats weight loss diet), (user has Eats high protein diet), \textbf{(user has Low fat/Low cholesterol diet)}, \textbf{(user has Low carbohydrate diet)}, \textbf{(Low fat/Low cholesterol diet match low\_cholesterol)}, \textbf{(Low fat/Low cholesterol diet match low\_saturated\_fat)}, \textbf{(Low carbohydrate diet match low\_carb)}

\vspace{5pt}

\textit{[Note prompt]} Important Note: Your output must be a Yes or No followed by strictly a list of nutrients with high or low as prefix among these options: carb, protein, sugar, sodium, cholesterol, saturated fat, calorie. For example, the output is: Yes, because the food is high in carb, low in protein, high in sugar. \\ \midrule
\textbf{True Answer} & Yes, because the food is low in carb, low in cholesterol, low in saturated\_fat. \\ \midrule
\textbf{LLM Answer} & \textcolor{ForestGreen}{Yes}, because the food is \textcolor{ForestGreen}{low in carb, low in cholesterol, low in saturated fat}. \\ \midrule
\textbf{LLM Reasoning} & 
The user has a Low fat/Low cholesterol diet and a Low carbohydrate diet. The given food, Vegetable rice soup, matches the user's dietary requirements as it belongs to low\_cholesterol, low\_saturated\_fat, and low\_carb categories. Therefore, the food is suitable for the user's needs. \\ \midrule
\textbf{Observation} & ToG’s retrieved subgraph contains all the necessary triples, no more no less, to determine that the food is healthy to the user’s diet requirements. \\
\bottomrule
\end{tabular}
\caption{Case Study 5}
\label{tab:case_5}
\end{table*}


\begin{table*}[t]
\centering
\small
\renewcommand{\arraystretch}{1.5}
\begin{tabular}{lp{12cm}}
\toprule
\textbf{Configurations} & Task: Text Generation \\
                        & Question: Complex \\
                        & Model: Llama-3.1-70B-instruct \\
                        & Method: ToG \\ \midrule
\textbf{Prompt} & 
\textit{[Question]} Based on the nutrients the food provides and the user needs, please answer whether the food \textbf{"Pie, custard, individual size or tart"} is healthy for the user? Please answer with a short sentence explaining why.

\vspace{5pt}

\textit{[Method prompt]} Below are the extra information you use to answer the question, note that you should not use your general knowledge and the answer is among this information. You will be given the textual description of a directed graph.

\vspace{5pt}

\textit{[Textualized graph]} \textit{Note: Bolded triples are those that appear in ToG’s retrieved subgraph.}

\vspace{3pt}

(Pie, custard, individual size or tart belongs to Cakes and pies), (Pie, custard, individual size or tart has Sugars, granulated), (Pie, custard, individual size or tart has Vanilla extract, imitation, no alcohol), (Pie, custard, individual size or tart has Cornstarch), (Pie, custard, individual size or tart has Egg, whole, raw, fresh), (Pie, custard, individual size or tart has Wheat flour, white, all-purpose, enriched, bleached), (Pie, custard, individual size or tart has Shortening, vegetable, household, composite), (Pie, custard, individual size or tart has Salt, table, iodized), (Pie, custard, individual size or tart has Milk, nonfat, fluid, without added vitamin A and vitamin D (fat free or skim)), (Pie, custard, individual size or tart has Beverages, water, tap, municipal), \textbf{(Pie, custard, individual size or tart belongs to low\_carb)}, (Pie, custard, individual size or tart belongs to high\_sodium), \textbf{(Pie, custard, individual size or tart belongs to low\_protein)}, \textbf{(Pie, custard, individual size or tart belongs to high\_cholesterol)}, \textbf{(Pie, custard, individual size or tart belongs to low\_saturated\_fat)}, (user has Drinks lots of milk), (user has Uses tobacco often), (user has Drinks Alcohol more than average), (user has Rarely or never check nutrition labels), (user has Takes few or no supplements), (user has Uses lots of salt in preparation), (user has Claims to have a poor diet), (user has Ate more food than usual), (user has Ate less food than usual), (user has Eats weight loss diet), (user has obesity), \textbf{(user has opioid\_misuse)}, (user has hypertension), \textbf{(user has diabetes)}, \textbf{(user has Low fat/Low cholesterol diet)}, (obesity need low\_calorie), \textbf{(opioid\_misuse contradict low\_protein)}, (opioid\_misuse need low\_sugar), (opioid\_misuse contradict high\_sodium), (hypertension contradict high\_sodium), (diabetes need low\_sugar), \textbf{(diabetes match low\_carb)}, \textbf{(Low fat/Low cholesterol diet contradict high\_cholesterol)}, \textbf{(Low fat/Low cholesterol diet match low\_saturated\_fat)}

\vspace{5pt}

\textit{[Note prompt]} Important Note: Your output must be a Yes or No followed by strictly a list of nutrients with high or low as prefix among these options: carb, protein, sugar, sodium, cholesterol, saturated fat, calorie. For example, the output is: Yes, because the food is high in carb, low in protein, high in sugar. \\ \midrule
\textbf{True Answer} & No, because the food is high in sodium, high in cholesterol, low in protein. \\ \midrule
\textbf{LLM Answer} & No, because the food is \textcolor{red}{low in carb}, \textcolor{ForestGreen}{low in protein, high in cholesterol}. \\ \midrule
\textbf{LLM Reasoning} & \textcolor{red}{The user has diabetes, and a low-carb diet is suitable for diabetes management. Therefore, the food being low in carb is a positive aspect}. However, the user also has opioid misuse, which contradicts with the food being low in protein. Protein is essential for overall health and recovery. Additionally, the user follows a Low fat/Low cholesterol diet. The food being high in cholesterol contradicts this requirement. \\ \midrule
\textbf{Observation} & ToG’s retrieved subgraph doesn’t contain the important triple “(hypertension contradict high\_sodium)”, hence it fails to include this information in the reasoning process. Instead, it incorrectly includes the food tag low\_carb into the final answer, despite successfully identifying that this is a positive aspect of the food. \\
\bottomrule
\end{tabular}
\caption{Case Study 6}
\label{tab:case_6}
\end{table*}


\begin{table*}[t]
\centering
\small
\renewcommand{\arraystretch}{1.5}
\begin{tabular}{lp{12cm}}
\toprule
\textbf{Configurations} & Task: Multi-label Classification \\
                        & Question: Complex \\
                        & Model: Llama-3.1-70B-instruct \\
                        & Method: ToG \\ \midrule
\textbf{Prompt} & 
\textit{[Question]} Based on the nutrients the food provides and the user needs, please answer what nutrient tags are used to determine whether the food \textbf{"Lasagna with cheese and meat sauce, reduced fat and sodium (diet frozen meal)"} is healthy or unhealthy for the user?

\vspace{5pt}

\textit{[Method prompt]} Below are the extra information you use to answer the question, note that you should not use your general knowledge and the answer is among this information. You will be given the textual description of a directed graph.

\vspace{5pt}

\textit{[Textualized graph]} \textit{Note: Bolded triples are those that appear in ToG’s retrieved subgraph.}

\vspace{3pt}

(Lasagna with cheese and meat sauce, reduced fat and sodium (diet frozen meal) belongs to Pasta mixed dishes, excludes macaroni and cheese), \textbf{(Lasagna with cheese and meat sauce, reduced fat and sodium (diet frozen meal) belongs to low\_carb)}, \textbf{(Lasagna with cheese and meat sauce, reduced fat and sodium (diet frozen meal) belongs to low\_sugar)}, \textbf{(Lasagna with cheese and meat sauce, reduced fat and sodium (diet frozen meal) belongs to high\_sodium)}, (Lasagna with cheese and meat sauce, reduced fat and sodium (diet frozen meal) belongs to low\_protein), \textbf{(Lasagna with cheese and meat sauce, reduced fat and sodium (diet frozen meal) belongs to low\_cholesterol)}, (Lasagna with cheese and meat sauce, reduced fat and sodium (diet frozen meal) belongs to low\_saturated\_fat), (user has Eats lots of shellfish), (user has Eats lots of fish), (user has Eats little to no frozen food), (user has Eats few to no meals outside home), (user has Eats few to no ready to eat meals), (user has Takes few or no supplements), (user has Uses lots of salt in preparation), (user has Claims to have a poor diet), (user has Ate more food than usual), (user has Ate less food than usual), (user has Eats weight loss diet), \textbf{(user has obesity)}, (user has opioid\_misuse), \textbf{(user has hypertension)}, \textbf{(user has diabetes)}, \textbf{(user has Low fat/Low cholesterol diet)}, (obesity need low\_calorie), (opioid\_misuse contradict low\_protein), (opioid\_misuse match low\_sugar), (opioid\_misuse contradict high\_sodium), \textbf{(hypertension contradict high\_sodium)}, \textbf{(diabetes match low\_sugar)}, \textbf{(diabetes match low\_carb)}, \textbf{(Low fat/Low cholesterol diet match low\_cholesterol)}, (Low fat/Low cholesterol diet match low\_saturated\_fat)

\vspace{5pt}

\textit{[Note prompt]} Important Note: Your output must be separated by comma, a list of nutrients with high or low before the nutrients among these options: carb, protein, sugar, sodium, cholesterol, saturated\_fat, calorie. For example, the output is: high\_carb, low\_protein, high\_sugar. Provide reasonings for your output. \\ \midrule
\textbf{True Answer} & low\_carb, low\_sugar, high\_sodium, low\_cholesterol, low\_saturated\_fat, low\_protein \\ \midrule
\textbf{LLM Answer} & \textcolor{ForestGreen}{high\_sodium, low\_sugar, low\_cholesterol, low\_carb} \\ \midrule
\textbf{LLM Reasoning} & 
- high\_sodium: Although the meal is labeled as "reduced sodium", it still belongs to the high\_sodium category. This is a concern for the user who has hypertension, as high sodium intake can exacerbate this condition.
- low\_sugar: The meal belongs to the low\_sugar category, which is good for the user who has opioid\_misuse.
- low\_cholesterol: The user has Low fat/Low cholesterol diet, which matches low\_cholesterol of the food.
- low\_carb: The food is low in carb, which is suitable for the user who has diabetes. \\ \midrule
\textbf{Observation} & The model is able to correctly identify 4 out of 6 important tags in this case. Although it is missing half of the tags, one good thing to note is that ToG is able to pick out both the matches (low\_cholesterol, low\_sugar, low\_carb) and the contradictions (high\_sodium). \\
\bottomrule
\end{tabular}
\caption{Case Study 7}
\label{tab:case_7}
\end{table*}

\section{Addtional Error Analysis}
\label{appendix:Error Analysis}

Our experiments showed that in the specific task of health-aware nutrition reasoning, LLMs are prone to two main types of errors: contextual hallucination and factual hallucination. To understand these shortcomings, we perform an error analysis focusing on the Text Quality Evaluation task, using 3 methods (KAPING, CoT-Zero, ToG) as a representative setting. We prompt the models to also include the reasonings behind their final answer, which then go through a human review process, revealing 2 types of reasoning failures: Contextual Hallucination and Factual Hallucination. Note that we do not check for KG topology errors, as our KG generation process ensures there are no structural problems in the knowledge base that would affect the model's information retrieval and processing performance. Exemplary demonstrations of these 2 error types are shown in Table-\ref{tab:error_analysis_1} and Table-\ref{tab:error_analysis_2}.

\begin{table*}[t]
\centering
\small
\renewcommand{\arraystretch}{1.5}
\begin{tabular}{lp{12cm}}
\toprule
\textbf{Question} & Based on the nutrients the food provides and the user needs, please answer whether the food \textbf{"Cheese pastry puffs"} is healthy for the user? Please answer with a short sentence explaining why. \\ \midrule
\textbf{User Profile} & Eats lots of shellfish, Eats lots of fish, Light cigarette smoker, Drinks alcohol less than average, Eats little to no fast food, Eats little to no frozen food, Often checks nutrition labels, Takes more supplements, Uses little to no salt in preparation, Drinks little water, Eats weight-gain diet, Has hypertension \\ \midrule
\textbf{Food Profile} & low\_carb, low\_sugar, high\_sodium, high\_calorie, low\_protein, high\_cholesterol, high\_saturated\_fat \\ \midrule
\textbf{True Answer} & No, because the food is high in sodium. \\ \midrule
\textbf{KAPING} & 
- \textbf{Answer:} No, because the food is high in sodium, high in calorie, high in saturated fat, high in cholesterol. \\
& - \textbf{Reasoning:} \textcolor{red}{Although the user often checks nutrition labels and eats little to no fast food or frozen meals} \textit{(factual hallucination - irrelevant information)}, their hypertension contradicts with the high sodium content found in Cheese pastry puffs. \textcolor{red}{Moreover, being a light smoker might increase cardiovascular risks; consuming foods high in calorie, saturated fat, and cholesterol like Cheese pastry puffs will not help.} \textit{(contextual hallucination - unaware of user's needs)} \\ \midrule
\textbf{CoT-Zero} & 
- \textbf{Answer:} No, because the food is high in sodium, low in protein, high in cholesterol, high in saturated fat, high in calorie.\\
& - \textbf{Reasoning:} The presence of 'high\_sodium' category contradicts with 'hypertension', as excessive sodium can worsen hypertension. \textcolor{red}{High saturated fat, calorie and cholesterol from Cheese and Margarine are also concerns for heart health. Low protein likely won't support muscle maintenance or satisfaction between meals.} \textit{(factual hallucination - irrelevant information)} \\ \midrule
\textbf{ToG} & 
- \textbf{Answer:} No, because the food is high in sodium. \\
& - \textbf{Reasoning:} \textcolor{ForestGreen}{Cheese pastry puffs is not healthy because it is high in sodium, and the user has hypertension.} \\ \bottomrule
\end{tabular}
\caption{Error Analysis 1}
\label{tab:error_analysis_1}
\end{table*}

\begin{table*}[t]
\centering
\small
\renewcommand{\arraystretch}{1.5}
\begin{tabular}{lp{12cm}}
\toprule
\textbf{Question} & Based on the nutrients the food provides and the user needs, please answer whether the food \textbf{"Shepherd's pie with lamb"} is healthy for the user? Please answer with a short sentence explaining why. \\ \midrule
\textbf{User Profile} & Drinks lots of milk, Eats little to no shellfish, Eats little to no fish, Drinks alcohol more than average, Eats lots of fast food, Eats few to no ready-to-eat meals, Rarely or never checks nutrition labels, Takes more supplements, Claims to have a poor diet, Has obesity, Has diabetes. \\ \midrule
\textbf{Food Profile} & low\_carb, low\_sugar, high\_sodium, low\_protein, low\_cholesterol, low\_saturated\_fat \\ \midrule
\textbf{True Answer} & Yes, because the food is low in carb, low in sugar. \\ \midrule
\textbf{KAPING} & 
- \textbf{Answer:} No, because the food is high in sodium, low in protein. \\
& - \textbf{Reasoning:} While Shepherd's pie with lamb contains some nutritious ingredients like vegetables and lean meat, \textcolor{red}{its high sodium content can be detrimental for someone with obesity and diabetes, as it may contribute to hypertension and other health issues. Additionally, the low protein content may not be sufficient for satiety and muscle maintenance.} \textit{(contextual \& factual hallucinations - using external knowledge to make assumptions about user's health \& include irrelevant information)} \\ \midrule
\textbf{CoT-Zero} & 
- \textbf{Answer:} No, because the food is high in sodium, low in cholesterol, low in saturated fat. \\
& - \textbf{Reasoning:} \textcolor{ForestGreen}{Given the user's obesity and diabetes, a diet low in calories, sugar, and carbohydrates is essential}. While Shepherd's pie with lamb is low in sugar and carbs, \textcolor{red}{its high sodium content may pose a risk for those with diabetes and obesity, making it less suitable for the user's health needs.} \textit{(contextual hallucination)} \\ \midrule
\textbf{ToG} & 
- \textbf{Answer:} Yes, because the food is low in sugar, low in carb. \\
& - \textbf{Reasoning:} \textcolor{ForestGreen}{Given that the user has diabetes, Shepherd’s pie with lamb is healthy because it is low\_sugar and low\_carb.} \\ \bottomrule
\end{tabular}
\caption{Error Analysis 2}
\label{tab:error_analysis_2}
\end{table*}

\section{Standards and Regulation}
\label{appendix:Standards and Regulation}

\begin{figure*}[htbp!]
	\centering
	\includegraphics[width=1\linewidth]{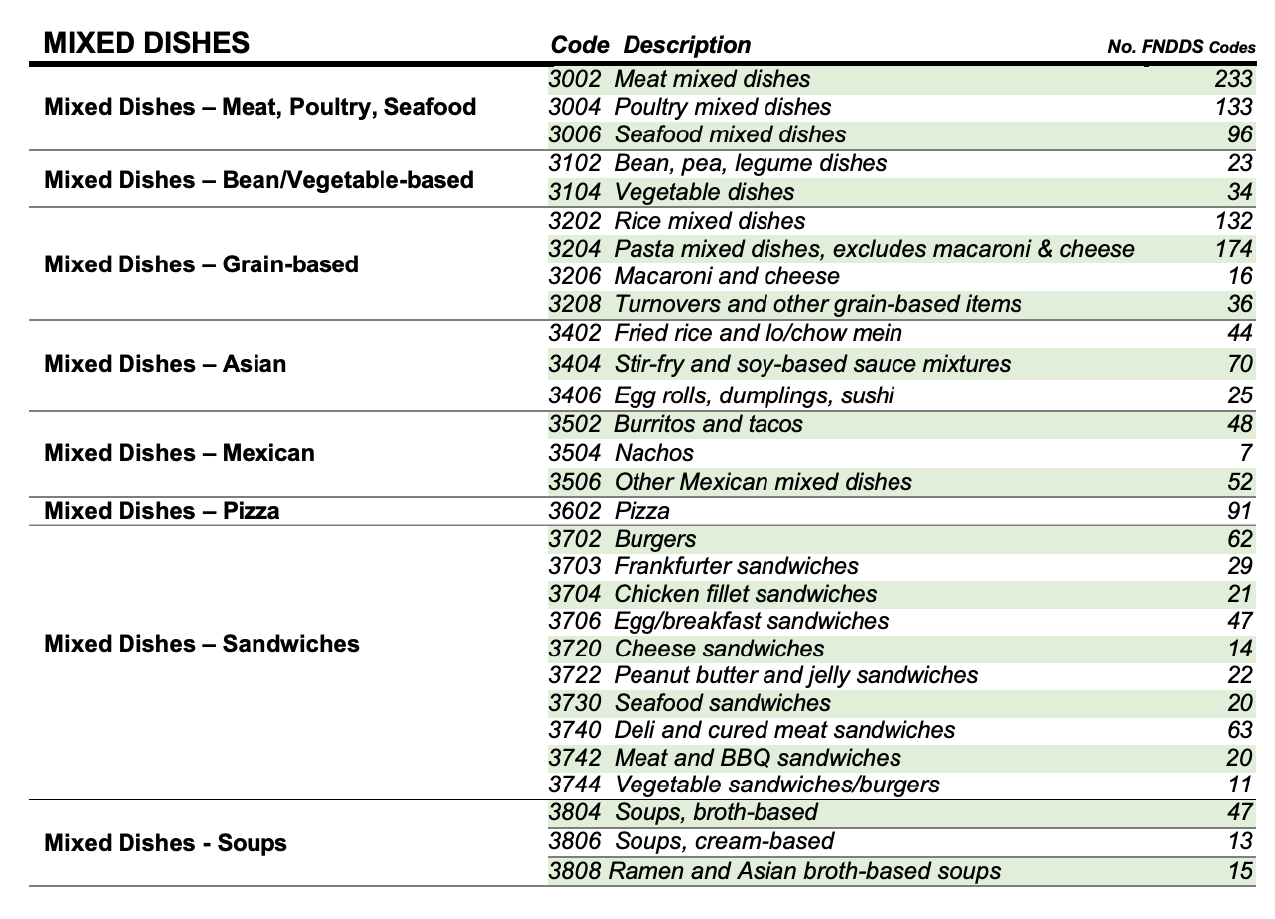}
        \vspace{-20pt}
	\caption{FNDDS Category Code - Mixed Dishes.}
        \vspace{-15pt}
    \label{fig:fndds}
\end{figure*}

\begin{figure*}[htbp!]
	\centering
	\includegraphics[width=1\linewidth]{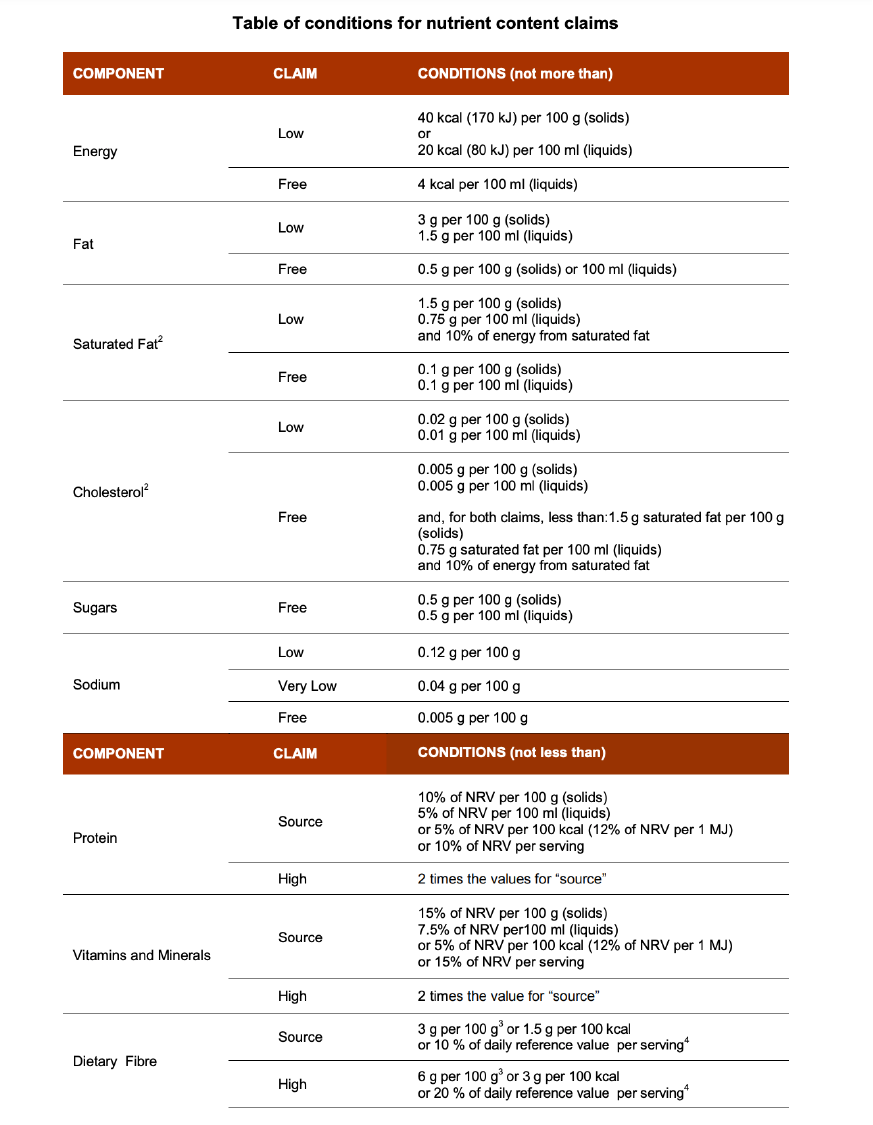}
        \vspace{-20pt}
	\caption{Guidelines for use of nutrition and health claims.}
        \vspace{-15pt}
    \label{fig:cac}
\end{figure*}

\begin{figure*}[htbp!]
	\centering
	\includegraphics[width=1\linewidth]{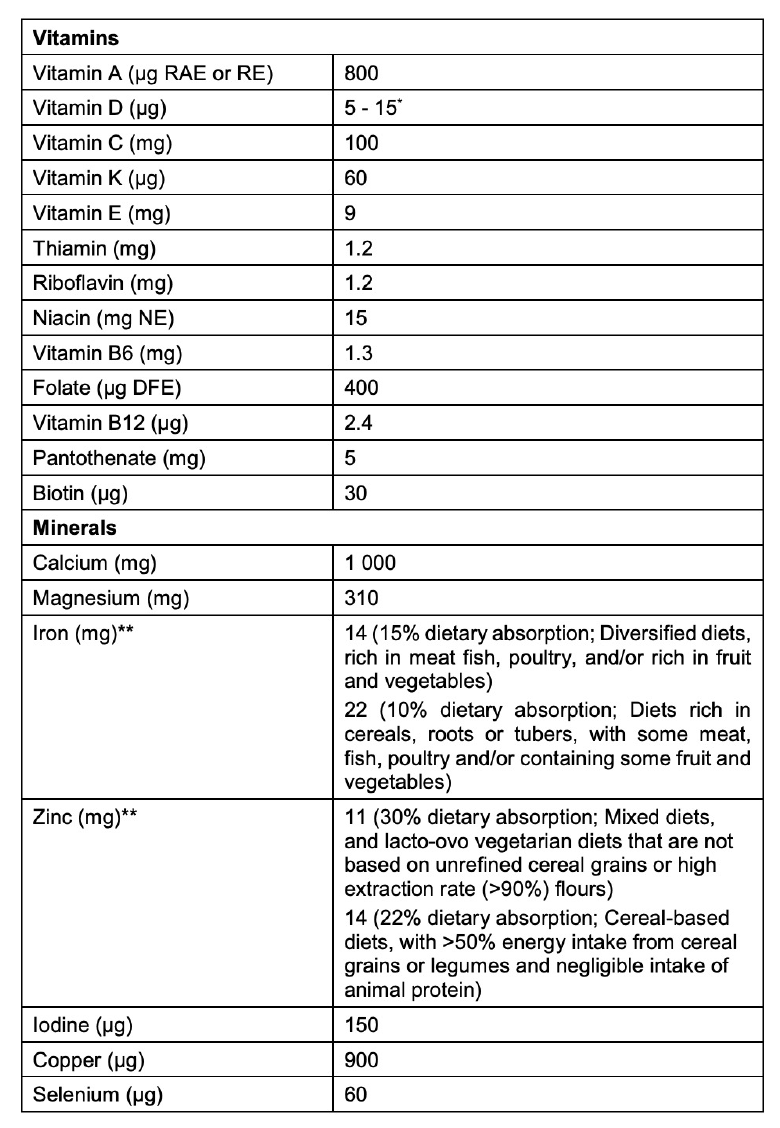}
        \vspace{-20pt}
	\caption{Daily nutrition value from Codex Alimentarius.}
        \vspace{-15pt}
    \label{fig:codex}
\end{figure*}

\begin{figure*}[htbp!]
	\centering
	\includegraphics[width=1\linewidth]{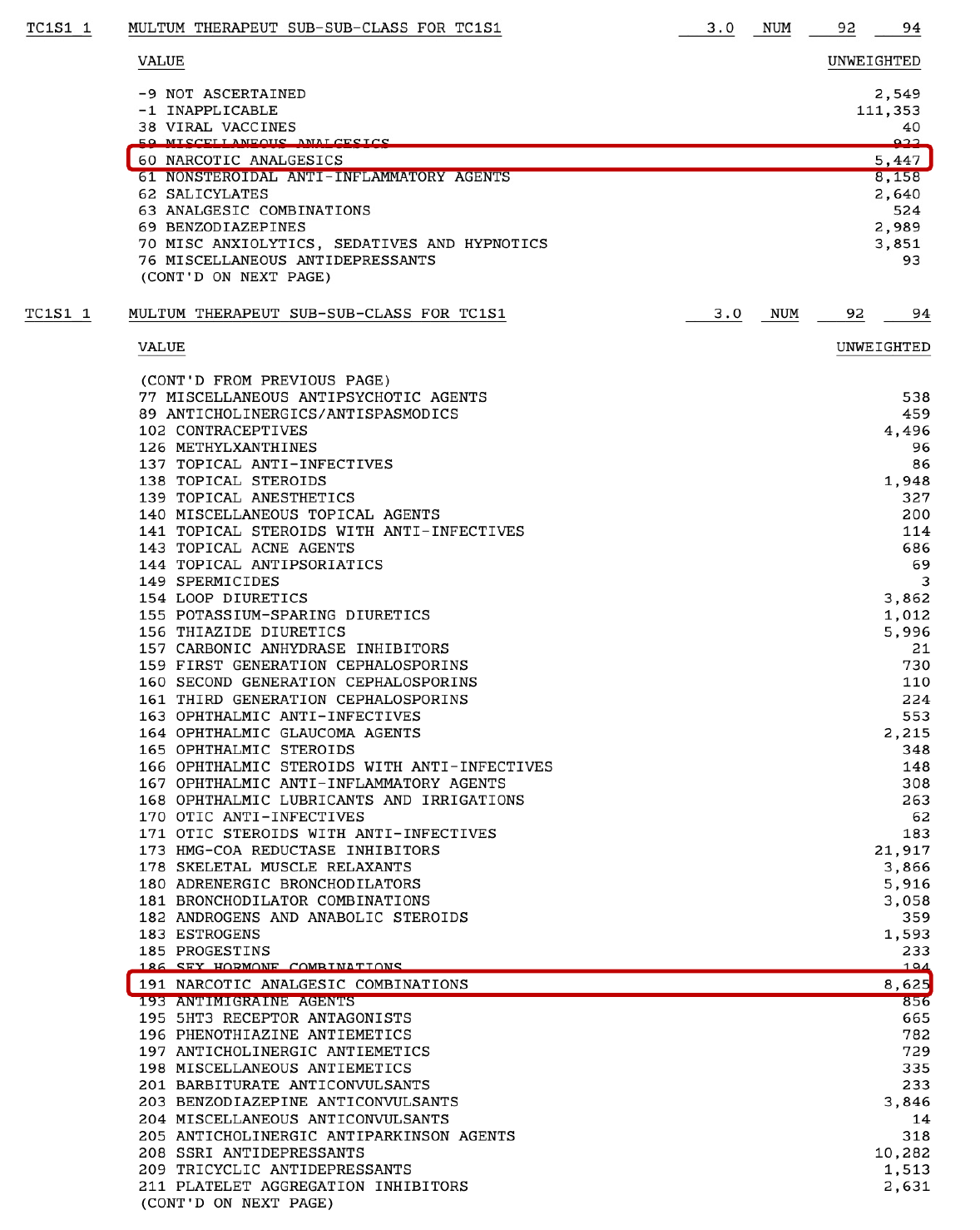}
        \vspace{-20pt}
	\caption{Multum Lexicon Therapeutic Classification Scheme - Part of Level 3.}
        \vspace{-15pt}
    \label{fig:lex}
\end{figure*}

In this section, we provide the standards and regulations used in this paper and attach their links of original document in footnote. There in general three categories: 1) The FNDDS category code \footnote{Full documention of FNDDS at \href{https://www.ars.usda.gov/ARSUserFiles/80400530/pdf/fndds/2021_2023_FNDDS_Doc.pdf}{here}} used for filtering food candidates (Figure-\ref{fig:fndds}). 2) Nutrition claim regulations from WHO, FSA\footnote{ \href{https://www.food.gov.uk/sites/default/files/media/document/fop-guidance_0.pdf}{FSA Guideline}}, CAC\footnote{\href{https://www.fao.org/fao-who-codexalimentarius/sh-proxy/en/?lnk=1&url=https\%253A\%252F\%252Fworkspace.fao.org\%252Fsites\%252Fcodex\%252FStandards\%252FCXG\%2B2-1985\%252FCXG_002e.pdf}{Guidelines on Nutrition Labeling}}\footnote{\href{https://www.fao.org/fao-who-codexalimentarius/sh-proxy/en/?lnk=1&url=https\%253A\%252F\%252Fworkspace.fao.org\%252Fsites\%252Fcodex\%252FStandards\%252FCXG\%2B23-1997\%252FCXG_023e.pdf}{Guidelines for Use of Nutrition and Health Claims}}, and EU legislation\footnote{\href{https://eur-lex.europa.eu/LexUriServ/LexUriServ.do?uri=OJ\%3AL\%3A2006\%3A404\%3A0009\%3A0025\%3AEn\%3APDF}{EU Nutrition \& Health Claims Regulation legislation (EC)}}.  used for defining nutrition thresholds (Figure-\ref{fig:cac} and Figure-\ref{fig:codex}) . Note that since there are discrepancies in the regulation. We adopt a stricter measure and make it sure it fits NHANES data. The Vitamins and Minerals high thresholds are calculated from the Daily Nutritional Reference Value (NRV), where CAC defines if a food (per 100g) contains over 15\% of NRV, it can claim itself a source of such nutrient. The Codex Alimentarius, or "Food Code" is a collection of standards, guidelines and codes of practice adopted by the Codex Alimentarius Commission. The Commission, also known as CAC, is the central part of the Joint FAO/WHO Food Standards Program and was established by FAO and WHO to protect consumer health and promote fair practices in food trade. 3) The Multum Lexicon Therapeutic Classification Scheme\footnote{Full document of Multum Lexicon Therapeutic Classification Scheme at \href{https://meps.ahrq.gov/data_stats/download_data/pufs/h68/h68f18cb.pdf}{here}}, used to define opioid prescription medicines and later mark opioid misuse (Figure-\ref{fig:lex}).

\end{document}